\pdfoutput=1

\documentclass[11pt]{article}
\usepackage[preprint]{acl}
\usepackage{rotating}
\usepackage{times}
\usepackage{latexsym}
\usepackage{multirow}
\usepackage[T1]{fontenc}
\usepackage{amsmath}
\usepackage[utf8]{inputenc}

\usepackage{microtype}

\usepackage{inconsolata}
\usepackage{booktabs} 
\usepackage{graphicx}
\usepackage{subfigure}
\usepackage{amsfonts}
\usepackage{amssymb}
\title{Mitigating the Language Mismatch and Repetition Issues in LLM-based Machine Translation via Model Editing}

\author{Weichuan Wang$^{\heartsuit, \clubsuit}$\thanks{~~The first two authors contributed equally to this work. They share the first authorship.},$\ $ Zhaoyi Li$^{\heartsuit, \spadesuit}$\textsuperscript{$*$}, Defu Lian$^{\spadesuit}$, Chen Ma$^{\heartsuit}$, Linqi Song$^{\heartsuit,\clubsuit \dag}$, Ying Wei$^{\diamondsuit}$\thanks{~~Corresponding authors.} \\
$^{\heartsuit}$City University of Hong Kong,
$^\spadesuit$University of Science and Technology of China\\
$^{\clubsuit}$City University of Hong Kong Shenzhen Research Institute, $^{\diamondsuit}$Zhejiang University\\
\texttt{weicwang2-c@my.cityu.edu.hk, lizhaoyi777@mail.ustc.edu.cn}\\ 
\texttt{liandefu@ustc.edu.cn, \{chenma,linqi.song\}@cityu.edu.hk, ying.wei@zju.edu.cn}
}

\begin{document}
\maketitle

\begin{abstract}
Large Language Models (LLMs) have recently revolutionized the NLP field, while they still fall short in some specific down-stream tasks. 
In the work, we focus on utilizing LLMs to perform machine translation, where we observe that two patterns of errors frequently occur and drastically affect the translation quality: language mismatch and repetition. 
The work sets out to explore the potential for mitigating these two issues by leveraging model editing methods, e.g., by locating Feed-Forward Network (FFN) neurons or something that are responsible for the errors and deactivating them in the inference time.
We find that directly applying such methods either limited effect on the targeted errors or has significant negative side-effect on the general translation quality, indicating that the located components may also be crucial for ensuring machine translation with LLMs on the rails.
To this end, we propose to refine the located components by fetching the intersection of the locating results under different language settings, filtering out the aforementioned information that is irrelevant to targeted errors. 
The experiment results empirically demonstrate that our methods can effectively reduce the language mismatch and repetition ratios and meanwhile enhance or keep the general translation quality in most cases.
\end{abstract}
\section{Introduction}

\begin{figure}[t]
\vspace{-0.1in}
    \centering
    \includegraphics[width=0.45\textwidth]{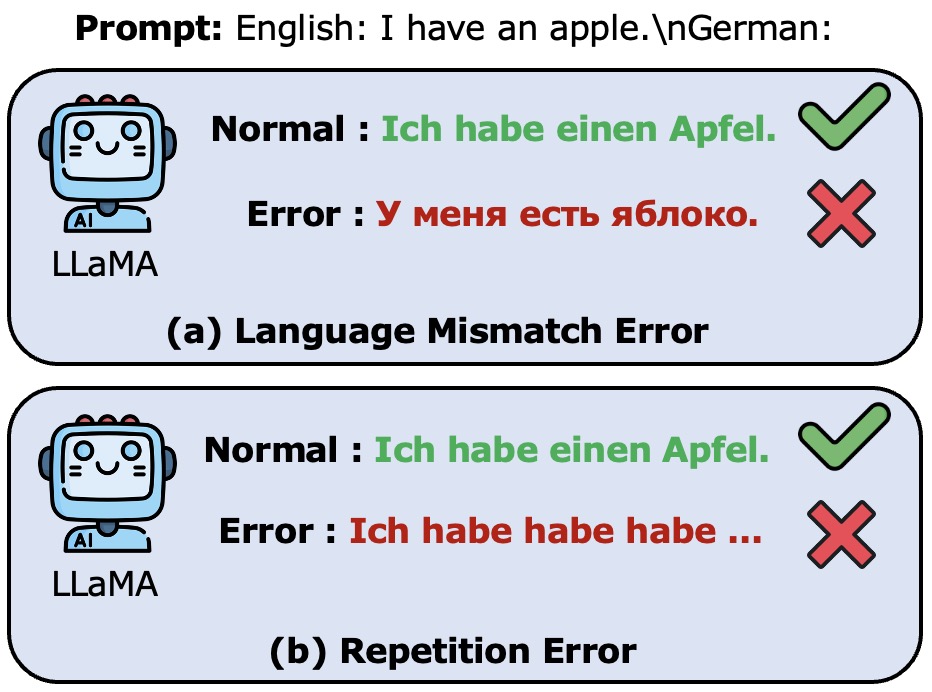}
    \caption{The illustration of the language mismatch error (a) and the repetition error (b).}
    \label{fig: error_illustrate}
\vspace{-0.1in}
\end{figure}

Pre-trained Large Language Models (LLMs) are natural machine translators with in-context learning \cite{gpt3, llama2, palmmt, bloommt, prompt-llmmt}, while they still fall behind specialized Machine Translation (MT) systems like NLLB \cite{nllb200}. Previous studies utilize In-Context Learning~ \cite{in-contextllmmt} (ICL), instruction tuning \cite{parashift, few-shottuningllm-mt} and post-editing methods \cite{parrotllm-mt, postllmmt, postllm-chat} to improve the translation quality. One further question is: \textit{Are there any specific issues that were ignored in previous studies hindering the LLM-based machine translation from further development?} In this work, we identify two issues in the LLM-based machine translation: \textit{Language Mismatch} and \textit{Repetition} (as shown in Figure~\ref{fig: error_illustrate}). We check the occurrence of these errors and find that: (1) they are common errors in the whole translation set (e.g., in the en$\rightarrow$de setting, language mismatch occurs in over $40\%$ cases with Zero-Shot prompting); (2) they are severe errors for machine translation systems (e.g., repetition errors usually lead to an over $50\%$ BLEU decrease compared with a standard generation). 

Nonetheless, the inherent reason for these errors still remains unclear, let alone patching them. In recent research works on model editing~ \cite{dai-etal-2022-knowledge,rome,fv}, they typically leverage analyzing tools like causal mediation analysis~ \cite{pearl2014interpretation,vig2020investigating}, integrated gradient attribution~ \cite{10.5555/3305890.3306024} to locate important component units (e.g., Feed-Forword Network (FFN) neurons, attention heads and stuff) that are highly responsible for specific behavior patterns of LLMs, and then precisely control these behaviors by manipulating the located components (e.g., amplifying or suppressing the activation values of neurons).
Inspired by these works, we ask a research question: \textit{Can we leverage model editing methods to mitigate aforementioned language mismatch and repetition issues?} 

To explore the potential of model editing on mitigating these errors, we set out to adapt two widely-used model editing techniques, Function Vectors~ \cite{fv} (FV) and Knowledge Neurons~ \cite{dai-etal-2022-knowledge} (KN), to MT scenarios in an aim to locate error-relevant component units inside LLMs.  
However, our empirical results show that directly adapting FVs and KNs either has limited effect on the targeted errors or has significant side-effect on the general translation quality, which indicates that the located component units may be not only responsible for targeted error patterns but also crucial for ensuring machine translation with LLMs on the rails and hence directly manipulate them could result in affecting the general translation behavior.

We then aim to filter out the error-irrelevant components from the located results. 
A possible hypothesis is that \textit{the location for the important error-relevant modules is supposed to be independent of translation language settings.}
After comparing the locating results under the different translation language settings (de$\rightarrow$en, en$\rightarrow$de, zh$\rightarrow$en and en$\rightarrow$zh), we do observe that a proportion of located component units are shared across different language settings, which valid that the error-related components are highly corresponding to the MT rather than individual languages. 
Grounded on this observation, we propose to refine the located components by fetching the intersection of the locating results under different language settings. 
The empirical results across different language settings demonstrate that the modified methods can effectively reduce the language mismatch and repetition ratios and meanwhile keep or enhance the general translation quality in most cases.

Our main contributions are three-fold:
\begin{itemize}
    \item We identify two patterns of errors in LLM-based MT that frequently occur and badly affect the translation quality: language mismatch and repetition.
    \item We investigate the potential for leveraging model editing methods (FV and KN) to reduce these errors. We find that directly adapting the editing methods either has limited effect on the targeted errors or has significant side-effect on the general translation quality.
    \item We propose to refine the located modules by fetching the intersection of the locating results under different language settings. We show that with the refined locating results we could arouse the potential for editing methods to handle the language mismatch and repetition errors and meanwhile enhance or keep the general translation quality in most cases. 
\end{itemize}

Additionally, The performance of our methods could sometimes be comparable with traditional methods that adapt LLMs to MT tasks (e.g., 5-Shot ICL \cite{in-contextllmmt}, LoRA \cite{lora} and Full-FineTuning \cite{sftsurvey}) without additional requirements like long-context prompting or fine-tuning. Besides, the proposed methods are compatible with the above techniques for further improvements.

\section{Related Work}
\textbf{Large Language Models for Machine Translation} 
One surprising ability of LLMs is that they are natural machine translators with Zero-Shot or One-Shot prompt \cite{gpt3, llama2, palmmt,bloommt,chatgptmt, prompt-llmmt}. However, there is still a gap \cite{parashift} between pre-trained LLM and large-scale NMT systems like NLLB \cite{nllb200} on the machine translation task. To bridge this gap, previous studies utilize in-context learning \cite{adaptivemt, in-contextllmmt, bloommt, palmmt}, model tuning \cite{parashift,few-shottuningllm-mt, baylingllmmt}, and interaction with annotation methods \cite{parrotllm-mt, postllmmt} to improve the translation quality. Even though LLM has achieved massive success in machine translation \cite{wmt23mt}, some of the issues from LLM itself may challenge machine translation, such as Hallucination \cite{hullucinationllm-mt}. Meanwhile, these problems from LLM are challenging to detect only with MT metrics.  \citet{few-shottuningllm-mt} find few-shot tuning can improve the translation quality based on MT metrics \cite{papineni-etal-2002-bleu, rei-etal-2022-comet} but detect the machine translation hallucination with a case-based hallucination design. In this work, we detect language mismatch and repetition issues
 in current LLM-based MT works, which are also found and regarded as
errors or hallucinations by some of previous works \cite{bloommt,few-shottuningllm-mt} but on a case study view.\\
\textbf{Locating Based Model Editing}
Precisely locating a small set of important modules (e.g., neurons~ \cite{dai-etal-2022-knowledge}, hidden states~ \cite{fv}, Multi-Head Self-Attention (MHSA)~ \cite{li2024understanding} and MLP~ \cite{rome} outputs) and editing their values to steer large-scale models toward assumed behaviours (e.g., updating factual associations~ \cite{rome,hase2023does}, detoxifying~ \cite{wang2024detoxifying}, decreasing hallucination~ \cite{li2023inferencetime}, switching languages~ \cite{tang2024languagespecific}, alleviating catastrophic forgetting~ \cite{jiang2024interpretablecatastrophicforgettinglarge} and patching reasoning errors~ \cite{li2024understanding}) is a recently emerging paradigm.
Nonetheless, such techniques are still largely under-explored in the context of MT.
In this work, we investigate the potential for adapting two representative locating-based editing approaches (specifically, Function Vectors~ \cite{fv} and Knowledge Neurons~ \cite{dai-etal-2022-knowledge}) to the MT scenario to mitigate its two fundamental but crucial issues: language mismatch and repetition~ \cite{zhang-etal-2021-generic}.

\section{Preliminary}
\label{sec:preliminary}
\label{sec:data}
In this section, we detail the data preparation process, including the data source, prompt template, and dataset construction. Additionally, we provide information about the model, the evaluation metrics used to support the ensuing experiments and the model editing methods used in this work.
\paragraph{Data Source} We choose three high-resource languages: \textit{English, Chinese, German} which show good performance on MT tasks \cite{chatgptmt}. For the detailed language setting, we include two language pairs: English-Chinese and English-German, and four translation directions: en$\rightarrow$de, de$\rightarrow$en, en$\rightarrow$zh and zh$\rightarrow$en (where en, de, zh represent English, German and Chinese, respectively). In the data choice, we use the human-made dataset from general MT tasks of WMT21, WMT22 and WMT23 \footnote{https://github.com/wmt-conference/wmt\textbf{X}-news-systems. \textbf{X}$\in\{21,22,23\}$} to ensure both high data quality and flexible data domain. These data make the machine translation approach a real-life usage to help us understand the current state of machine translation using LLMs.
\paragraph{Prompt Template} For machine translation tasks, a widely-adopted~ \cite{prompt-llmmt, bloommt, palmmt} K-Shot In-Context Learning (ICL) prompt template (taking the language setting of en$\rightarrow$zh for an example) is:
\begin{equation}
\begin{split}
&\text{English}:src_1\backslash n\text{Chinese}:tgt_1\backslash n\\
&...\\
&\text{English}:src_K\backslash n\text{Chinese}:tgt_K\backslash n\\
&\text{English}:src_q\backslash n\text{Chinese}:\nonumber
\end{split}
\end{equation}
Where $(src_i,tgt_i)$ refers to the $i$-th in-context translation exemplar ($src_i$ refers to a sentence of source language and $tgt_i$ refers to the corresponding sentence of target language.). $src_q$ refers to the real sentecne of source language that needs to be translated.
We call this prompt template \textit{Lang Prompt} and regard it as the default prompt template for the follow-up experiments in this paper.
\paragraph{Dataset Construction} 
In the data construction part, we construct the $\mathcal{D}_{exps}$ (data from WMT21) to provide the ICL exemplars used in the K-Shot prompt for machine translation tasks. We use the WMT22 data as the $\mathcal{D}_{train}$ to fine-tune a model or locate the crucial parts in an LLM for model editing methods. For the testing and validation, we construct the $\mathcal{D}_{test}$ (data from WMT23) for various modifications (e.g. fine-tuning \cite{finetune} or model editing methods \cite{fv,dai-etal-2022-knowledge}). (Please refer to Appendix ~\ref{app:details_data_construct} for detailed dataset information)
\paragraph{Model} To support the in-depth exploration and analysis of how the two kinds of errors happen. We use LLaMA2-7B as our backbone language model to implement the machine translation task and further adaptation \cite{llama2}. (We also explore the scaling experiments on LLaMA2-13B with the same data and methods, which can be seen in Appendix \ref{scaling_exps})
\paragraph{Evaluation Metrics}
For the machine translation metrics, we consider the overlapping-based metrics BLEU \cite{papineni-etal-2002-bleu} and neural-based metrics COMET22DA \cite{rei-etal-2022-comet} to evaluate the translation quality (For a detailed toolkit and detection process, please refer to Appendix ~\ref{app:details_eval}).

\paragraph{Model Editing Methods}
For the concrete model editing methods, we choose Function Vectors (FV)~ \cite{fv} and Knowledge Neuron (KN)~ \cite{dai-etal-2022-knowledge}:
\textit{FV} argues that the key information of a task ($\mathcal{T}$) is compactly represented and transported in a small set of attention heads in LLMs. Then, they utilize the summation of these located head vectors and directly add the integrated vector to the "residual stream"~ \cite{elhage2021mathematical} of forwarding computation of Transformer-based~ \cite{transformer} LLMs to help them perform ideal behaviour of task $\mathcal{T}$.
\textit{KN} further develops the idea of viewing the Feed-Forward Networks (FFNs) in the Transformer~ \cite{transformer} as key-value memories~ \citet{geva-etal-2021-transformer} (memories can be specific words, specific topics and factual knowledge) and locating a small set of neurons in the FFNs that highly attribute to factual knowledge to manipulate.

\section{Language Mismatch and Repetition Error in LLM-MT}\label{sec:two_issues}
In our initial experiments, we observe that LLM-based machine translation struggles with the following two types of common errors.
One is \textbf{Language Mismatch}, referring to the language of the translation result is not the target language. 
For example, In the en$\rightarrow$zh machine translation, the target language is Chinese while the language of generated sentence is still English. 
Another is \textbf{Repetition}, 
referring to a substring is generated repeatedly until the end of the generation. 
To evaluate these errors, we additionally introduce two metrics: \textbf{L}anguage \textbf{M}ismatch \textbf{R}atio (LMR) (the percentage of cases occurring the language mismatch error) and \textbf{R}epetition \textbf{R}atio (RR) (the percentage of cases occurring the repetition error).

\paragraph{Language mismatch and repetition error are common and crucial}
After detecting these errors, we first try to provide a quantitive analysis by analyzing the ratio of language mismatch and repetition error in Zero-Shot and One-Shot. For detailed language settings, we consider en$\rightarrow$de, de$\rightarrow$en, en$\rightarrow$zh, and zh$\rightarrow$en. We utilize the $\mathcal{D}_{test}$ and $\mathcal{D}_{exps}$ as the test set and prompt examplar source, respectively. We choose LMR and RR to represent the ratio of language mismatch and repetition in a setting (e.g. en$\rightarrow$de (Zero-Shot)). For translation quality evaluation, we choose the BLEU~ \cite{papineni-etal-2002-bleu} as the metrics since these errors can easily be detected on the word level with a sharp decrease on BLEU or human check. We observe that language mismatch is frequent in Zero-Shot and seldom in One-Shot. Repetition error cases in One-Shot are without language mismatch but combined with language mismatch in Zero-Shot. Based on our observation, we do experiments and analysis in Zero-Shot for the language mismatch and in One-Shot for the repetition error.

To explore the relation between the above errors and translation quality, we split the translation results into four sets to evaluate the BLEU performance after error detection. The four sets include two error sets: \textit{language mismatch set} and \textit{repetition error set}, one \textit{regular set} (where instances without both errors), and one \textit{Origin set} that includes all cases.
The results of Table~\ref{tab:general_error_wmt23_wmt22} illustrate: (1) the gap between the regular set and the original set shows both language mismatch and repetition error hurt the translation quality; (2) Language mismatch is the main reason for the low performance in Zero-Shot; (3) Even though we observe a low repetition ratio in One-Shot, the gap between repetition set and regular set shows that repetition is a severe error in the original set; (4) The performance gap between regular and error cases indicates a direct way to improve the translation quality by eliminating these errors. 

In this section, we run all experiments by using the \textit{Lang Prompt}~ \cite{prompt-llmmt, bloommt, palmmt} as the default prompt template. Currently, we notice that other prompt templates are used in LLM-based MT research~ \cite{bloommt, PaLMmodel, gpt3, xglm, CoTprompt}. To comprehensively explore these errors, we test other prompt templates with the same data and find these errors again. The only difference is the concrete ratio. This extension experiment further demonstrates the conclusion that \textit{language mismatch and repetition error are common and crucial}. (Detailed experimental setting and results can be found in Appendix~\ref{app_compre_prompts})
\begin{table}[ht]
\vspace{-0.1in}
\centering
\renewcommand{\arraystretch}{1}
\resizebox{\linewidth}{!}{
\begin{tabular}{lcccc}
\hline
Setting           & \textbf{L}($\downarrow$) & \textbf{OB}($\uparrow$) & \textbf{LB}($\uparrow$) & \textbf{RB}($\uparrow$) \\ \hline
zh$\rightarrow$en \textbf{(Z)} &0.0486   &17.13    &8.77   &17.60      \\
en$\rightarrow$zh \textbf{(Z)} &0.3269   &16.34    &3.13  &25.29    \\
en$\rightarrow$de \textbf{(Z)} &0.4524   &12.61    &1.65  &21.86     \\
de$\rightarrow$en \textbf{(Z)} &0.0219   &35.34    &23.23 &35.66      \\ \hline
Setting           & \textbf{R}($\downarrow$) & \textbf{OB}($\uparrow$) & \textbf{RRB}($\uparrow$) & \textbf{RB}($\uparrow$) \\ \hline \hline
zh$\rightarrow$en \textbf{(O)}  &0.0035   &18.87 &2.13 &19.06     \\
en$\rightarrow$zh \textbf{(O)}  &0.0146   &27.78 &2.08 &29.47 \\
en$\rightarrow$de \textbf{(O)}  &0.0141   &24.97 &12.64&25.86\\
de$\rightarrow$en \textbf{(O)}  &0.0018  &36.54  &6.10 &36.71  \\ \hline
\end{tabular}
}
\caption{The correlation between error ratio and BLEU. \textbf{(Z)} represents the Zero-Shot prompting, and \textbf{(O)} represents the One-Shot prompting. \textbf{L}: language mismatch ratio; \textbf{R}: repetition ratio; \textbf{OB}: The BLEU on the original set; \textbf{LB}: The BLEU on the language mismatch set; \textbf{RRB}: The BLEU on the repetition error set; \textbf{RB}: The BLEU on the regular set.}
\label{tab:general_error_wmt23_wmt22}
\vspace{-0.1in}
\end{table}

\section{Can we mitigate language mismatch and repetition via model editing?}
\label{sec:can_we_mitigate}
In this section, We aim to investigate the potential for leveraging model editing methods~ \cite{dai-etal-2022-knowledge,rome,fv} to precisely mitigate the aforementioned two severe issues in MT: language mismatch and repetition. We mainly focus on two widely-used model editing methods: \textbf{F}unction \textbf{V}ectors (\textbf{FV})~ \cite{fv} and \textbf{K}nowledge \textbf{N}eurons (\textbf{KN})~ \cite{dai-etal-2022-knowledge}, for both of them are representative (i.e., Causal Mediation Analysis~ \cite{rome,pearl2014interpretation} for FV and Integrated Gradient Attribution~ \cite{qi2019visualizing,lundstrom2022rigorous} for KN) and influential~ \cite{bai2024identifying,hojel2024finding,niu2024what,Chen_Cao_Chen_Liu_Zhao_2024}.
In the following paragraphs, we adapt the idea of FV (corresponding to \textbf{Machine translation vectors}) and KN (corresponding to \textbf{Machine translation neurons} and \textbf{Repetition neurons}) to MT scenarios, with an aim to both enhance the LLMs’ understanding (for both language mismatch and repetition errors) and ability on MT by handling these errors.
\subsection{Machine Translation Vectors}
FV has demonstrated that it can uncover partial mechanisms of some simplified human-designed tasks by adding a function vector of tasks. But what about a more complex and natural NLP task like MT?
To answer this difficult question, we begin with a direct and natural question: \textbf{\textit{Can we use FV to enhance LLMs’ understanding to MT and mitigate aforementioned language mismatch and repetition issues?}}
We use Ten-Shot ICL prompts $\mathcal{P}$ (the template of machine translation prompts is the \textit{Lang Prompt}~ \cite{prompt-llmmt} in Section~\ref{sec:preliminary}.) to locate important attention heads, where the data are sampled from $\mathcal{D}_{train}$.
For brevity, we denote the normal Ten-Shot ICL input (omitting language signs, i.e., “English”, “Chinese” and “German”) as: $inp = [(src_1, tgt_1),(src_2, tgt_2),...,(src_{10},tgt_{10}),src_{q}]$ $\in \mathcal{P}$, where $src$ and $tgt$ refer to sentences of source and target languages respectively; index $1\sim10$ refers to ten ICL exemplars and $q$ refers to “query” (the real source sentence that requires to be translated.). On its basis, we construct the \textit{shuffled} version of the original ICL input: $\widetilde{inp}=[(src_1, \widetilde{tgt_1}),(src_2, \widetilde{tgt_2}),...,(src_{10},\widetilde{tgt_{10}}),src_{q}]$, where for each ICL exemplar $(src_k, \widetilde{tgt_k})$, $k\in[1..10]$, the target sentence $\widetilde{tgt_k}\neq tgt_k$.
\paragraph{Extracting machine translation vectors} 
First, we locate attention heads that are important to the MT with a \textit{Causal Mediation} procedure: 
(1) Extract the average attention head output on Ten-Shot cases: $\overline{h}^i_j=\mathop{\mathbb{E}}\limits_{inp\in\mathcal{P}}[h^i_j(inp)]$, where $h^i_j$ means the $i$-th head of $j$-th layer.
(2) Send both $inp$ and $\widetilde{inp}$ to the same LLMs (denoting the model as $\theta$), 
(3) Fetch probabilities of predicting the ground-truth target sentence $tgt_{q}$ from models with the shaffled input: $p_{\theta}(tgt_{q}|\widetilde{inp})$,
(4) Adopt intervention: replacing \textit{a single attention head output} in the shuffled run with $\widetilde{inp}$ with the averaged attention head output extracted in step (1) at the same place ($h^i_j$), 
(5) Calculate the \textbf{C}ausal \textbf{I}ndirect \textbf{E}ffect ($\text{CIE}(\overline{h^i_j}\rightarrow h^i_j|inp)$) of the intervention on each Ten-Shot case: $p_\theta(tgt_q|\widetilde{inp}, \overline{h^i_j}\rightarrow h^i_j)-p_\theta(tgt_q|\widetilde{inp})$ and 
(6) Calculate the \textbf{A}verage \textbf{I}ndirect \textbf{E}ffect for head $h^i_j$: $\text{AIE}(h^i_j) = \mathop{\mathbb{E}}\limits_{inp\in\mathcal{P}}[\text{CIE}(\overline{h^i_j}\rightarrow h^i_j|inp)]$. 

\begin{figure}[!ht]
\vspace{-0.1in}
    \centering
    \subfigure[de$\rightarrow$en]{\includegraphics[width=0.49\linewidth]{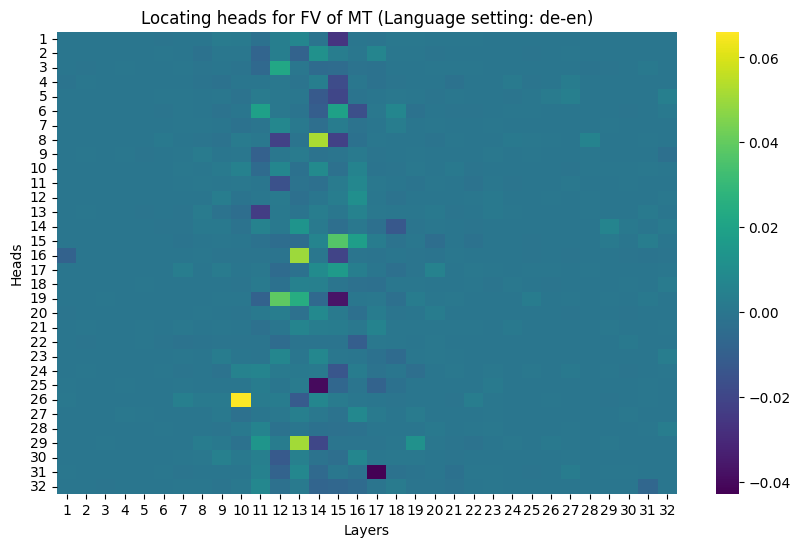}} 
    \subfigure[en$\rightarrow$zh]{\includegraphics[width=0.49\linewidth]{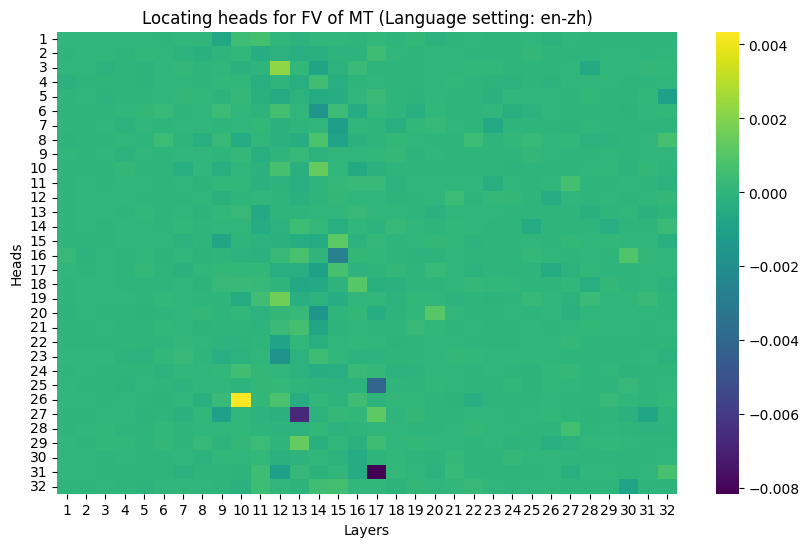}} 
\vspace{-0.1in}
    \caption{Heatmaps of AIE values for attention heads in LLaMA2-7B for de$\rightarrow$en setting (a) and en$\rightarrow$zh setting (b). x-axis and y-axis refer to the layer and head. Brighter color refers to the head with larger AIE value.}
    \label{fig:locating_head_main_text}
\vspace{-0.1in}
\end{figure}

The AIE values for all heads in LLaMA2-7B under the language settings\footnote{Due to the page limit, We post experiment results only under part of the language settings results in the main text. For the rest language settings, we post them in Appendix \textcolor{blue}{\ref{app:all_aies}}, Similarly hereinafter.} of “de$\rightarrow$en” and “en$\rightarrow$zh” are depicted in Figure\ref{fig:locating_head_main_text}. 
We observe that for machine translation there are sparsely a few heads of which the corresponding AIE values strikingly stand out among 1024 heads. We select top-32 heads (the number of heads in a layer and according to their AIE values, denoted as $\mathcal{H}$) to extract FV in the follow-up experiments.

Let $h^i_j(inp)$ denote the output of attention head $h^i_j$ given the input prompt $inp$. Following ~ \citet{fv}, we extract the machine translation vector with a specific language setting $\mathcal{V}_{\text{X}\rightarrow\text{Y}}$ (e.g., $\mathcal{V}_{\text{zh}\rightarrow\text{en}}$ means the language setting of zh$\rightarrow$en )  with the following formula:
\begin{align}
    \mathcal{V}_{\text{X}\rightarrow\text{Y}} = \mathop{\mathbb{E}}\limits_{inp\in\mathcal{P}_{\text{X}\rightarrow\text{Y}}}[\sum\limits_{h^i_j \in \mathcal{H}}h^i_j(inp)]
\end{align}
\paragraph{Editing LLMs via machine translation vectors} 
We directly add the extracted machine translation vector to the “residual stream” (being aligned with the original FV paper, at $11$-th layer for LLaMA2-7B) in the forwarding process. 
The performance of LLaMA2-7B (e.g., under the language setting of zh$\rightarrow$en.) after adopting machine translation vectors 
are posted in Table~\ref{tab:direct_adaptation_main_text}.

\begin{table}[t]
\vspace{-0.1in}
\centering
\resizebox{\linewidth}{!}{
\begin{tabular}{lccc}
\hline
\textit{\textbf{Zero-Shot}} & \textbf{L}($\downarrow$)  & \textbf{B}($\uparrow$) & \textbf{C}($\uparrow$)  \\
\hline                     
LLaMA2-7B  &$0.0486$&$17.1288$&$0.722$\\
+\textit{MT vectors}&$-72.84\%$&$-37.35\%$&$-1.84\%$\\                
+\textit{MT neurons} &$-18.72\%$&$-4.28\%$&$-0.15\%$\\
\hline
\hline
\textit{\textbf{One-Shot}} & \textbf{R}($\downarrow$)  & \textbf{B}($\uparrow$) & \textbf{C}($\uparrow$)  \\
\hline                      
LLaMA2-7B  &$0.0035$&$18.8714$&$0.7376$\\  
+\textit{MT vectors} &$482.86\%$&$-23.07\%$&$-1.68\%$\\                
+\textit{MT neurons} &$0.0\%$&$-0.35\%$&$-0.03\%$\\
+\textit{RP neurons} &$-8.57\%$&$0.07\%$&$0.0\%$\\
\hline
\end{tabular}
\vspace{-0.1in}
}
\caption{
Performance of LLaMA2-7B (and after applying model editing methods) on $\mathcal{D}_{test}$ (under the language setting of zh$\rightarrow$en). \textit{\textbf{Zero-Shot}} and \textit{\textbf{One-Shot}} refer to that using zero-shot prompt (for language mismatch errors) and one-shot prompt (for repetition errors) for MT. For evaluation metrics, \textbf{L}: Language mismatch ratio; \textbf{R}: Repetition ratio; \textbf{B}: BLEU and \textbf{C}: COMET22DA, where \textbf{B} and \textbf{C} mainly evaluate the general translation quality. For plain LLaMA2-7B, the results are absolute values; for LLaMA2-7B with editing methods, the results are relative \textbf{improvement percentages}.
}
\vspace{-0.1in}
\label{tab:direct_adaptation_main_text}
\end{table}
We observe that leveraging machine translation vectors (+\textit{MT vectors}) can
(1) reduce the language mismatch errors to a large extent ($-72.84\%$) while simultaneously
(2) introduce more repetition errors ($+482.86\%$) and
(3) do harm to the general translation quality: $-37.35\%$ (Zero-Shot) and $-23.07\%$ (One-Shot) for BLEU. (For the results of other language settings, we include them in Appendix~\ref{app_direct_adapt}.)
\subsection{Machine Translation Neurons and Repetition Neurons}
Beyond the original exploration of KN on factual knowledge, we also want to know the potential of KN on MT: \textbf{\textit{Can we use KN to locate and manipulate skilled neurons responsible for MT or the repetition error pattern?}}
In the MT scenarios, We denote the input prompt $inp$ (also omitting language sign) as $[src_q]$ (Zero-Shot) or $[(src_0,tgt_0),src_q]$ (One-Shot) and the corresponding output as $tgt_q$, where the $(src_0,tgt_0)$ is the ICL exemplar (sampled from $\mathcal{D}_{exps}$) and $(src_q,tgt_q)$ is the “query”, the real case used for locating neurons (sampled from $\mathcal{D}_{train}$) or testing edited models (sampled from $\mathcal{D}_{test}$).
\paragraph{Locating Important Neurons for MT}
We randomly sample a token $t$ in each $tgt_q$ (without errors) and use $t$ to split $tgt_q$ into two parts: $tgt_q=(\overset{\leftarrow}{tgt_{q}},\overset{\rightarrow}{tgt_{q}})$ ($t\in\overset{\rightarrow}{tgt_{q}}$). 
To fully model the MT and meanwhile restrict the computation, we focus on the probability of $p(t|inp^+)$, where $t$ refers to the first token of $\overset{\rightarrow}{tgt_{q}}$ and $inp^+$ refers to the concatenation of $inp$ and $\overset{\leftarrow}{tgt_{q}}$.
Focusing on a single neuron $w_i^{(l)}$ ($i$-th intermediate neuron in the $l$-th FFN), we denote its activation value as $\overline{w_i}^{(l)}$. Then we can introduce this variable into $p(t|inp^+)$ as $p(t|inp^+,w_i^{(l)}=\overline{w_i}^{(l)})\triangleq f(\overline{w_i}^{(l)})$ (fixing $t$ and $inp^+$, the probability can be viewed as an objective function whose only variable is the value of neuron $w_i^{(l)}$).
We calculate the attribution score of neuron $w_i^{(l)}$ by Integrated Gradient~ \cite{10.5555/3305890.3306024}: 
\begin{align}
\label{eq:attribution}
\text{Attr}(w_i^{(l)}|f)=\overline{w_i}^{(l)}\int_{\alpha=0}^{1} \frac{\partial f(\alpha\overline{w_i}^{(l)})}{\partial w_i^{(l)}} d\alpha.
\end{align}
We calculate the mean value of the attribution scores for each neuron with 2,000 examples through Riemann approximation with $20$ steps. We select top-5 neurons as Machine Translation neurons (\textit{MT neurons}).
\paragraph{Locating Important Neurons for Repetition}
We first collect all of examples that occur the repetition error. For a specific input prompt $inp$, the completed generation $y$ of a LLM can be divided into the following several parts: $y = [y_{norm},y_{repe},y_{repe},y_{rest}]$, where $y_{norm}$ refers to the normal generation part (except for the first-time generation of $y_{repe}$), $y_{repe}$ refers to the minimal repetition unit (the first $y_{repe}$ here is supposed to be treated as normal generation) and $y_{rest}$ (the follow-up generation after the second-time generation of $y_{repe}$).
To concentrate on the repetition error, we construct a new input prompt $inp_{repe}=[inp, y_{norm}, y_{repe}]$ and focus on the probability of $p(y_{repe}|inp_{repe})$. Similar to the \textit{MT neurons} part, we define neuron $w_i^{(l)}$, its value $\overline{w_i}^{(l)}$, its objective function $p(y_{repe}|inp_{repe},w_i^{(l)}=\overline{w_i}^{(l)})\triangleq f_{repe}(\overline{w_i}^{(l)})$ and its attribution score $\text{Attr}(w_i^{(l)}|f_{repe})$ (repetition attribution score).
A natural concern here is that \textbf{\textit{the objective function $f_{repe}(\overline{w_i}^{(l)})$ might model the pattern of generating $y_{repe}$ rather than the repetition error pattern}}.
To exclude this concern, we additionally set a comparison objective function $f_{compare} = p(y_{repe}|[inp,y_{norm}],w_i^{(l)}=\overline{w_i}^{(l)})$ to model the first-time generation (normal generation) of $y_{repe}$.
With $f_{compare}$, we can also get the attribution score $\text{Attr}(w_i^{(l)}|f_{compare})$ (comparison attribution score) of neuron $w_i^{(l)}$.
We calculate the mean values of repetition and comparison attribution scores separately for each neuron $w_i^{(l)}$ with all of the cases in $\mathcal{D}_{train}$ that occur the repetition error. 
We separately select top-300 neurons according to mean repetition and comparison attribution score, denoting the fetched sets as $\mathcal{N}_{repe}$ and $\mathcal{N}_{compare}$. We select 5 neurons with the largest repetition attribution scores from $\mathcal{N}_{repe}\backslash\mathcal{N}_{compare}$ as the Repetition Neurons (\textit{RP neurons}).
\paragraph{Editing LLMs via \textit{MT} neurons and \textit{RP} neurons}
For \textit{MT} neurons, we edit LLMs by \textit{amplifying} the activation values of these neurons (set the new values to be twice the original ones). 
For \textit{RP} neurons, we edit LLMs by \textit{erasing} the activation values of these neurons (set the new values to be zero).
The performance of LLaMA2-7B (e.g., under the language setting of zh$\rightarrow$en.) after adopting \textit{MT} neurons and \textit{RP} neurons are posted in Table~\ref{tab:direct_adaptation_main_text}.
We observe that 
(1) adopting \textit{MT} neurons can indeed help reduce language mismatch ratio to some extent($-18.72\%$) while also bring small negative side-effect to the translation quality ($-4.28\%$ for the BLEU score),
(2) adopting \textit{MT} neurons nearly have no effect on the repetition ratio and
(3) adopting \textit{RP} neurons can reduce the repetition ratio slightly ($-8.57\%$) without affecting the metrics (BLEU and COMET22DA) of evaluating general translation quality.

Hence a short response to the question of this section is that \textit{\textbf{Directly leveraging model editing methods either has limited effect on errors (MT neurons and RP neurons) or significant negative side-effect on general translation quality (MT vectors)}}. Nonetheless, we do observe the potential for mitigating the aforementioned errors with editing methods.

\section{Modifications to FV and KN in MT scenarios}
\label{sec:modifications}
In section, we mainly discuss our modifications (Section~\ref{sec:modification_methods}) to FV and KN methods (Section~\ref{sec:can_we_mitigate}) to release their potential for better mitigating the language mismatch errors, repetition errors and hopefully improving the general translation quality. Besides, we present systematical evaluation results for the modified editing methods and baselines in Section~\ref{sec:modification_results} to facilitate a deeper understanding of LLM-based MT.
\subsection{Modifications}
Previous empirical results (Section~\ref{sec:can_we_mitigate}) show that \textit{MT vectors} are more effective to reduce language mismatch errors in comparison with \textit{MT neurons} while the \textit{RP neurons} are more promising for handling repetition errors, suggesting that the inherent mechanisms for the recognition of target language and generating strings repeatedly locate in heads and FFN neurons of LLMs, respectively. 
To this end, in the follow-up experiments, we concentrate on modifying \textit{MT vectors} to handle language mismatch errors and \textit{RP neurons} to handle repetition errors.
Our first modification is based on a natural hypothesis: \textbf{\textit{The location for the important modules inside LLMs that are responsible for target language recognition and repetition errors is supposed to be independent to language settings.}}
\label{sec:modification_methods}
\begin{figure}
\vspace{-0.1in}
    \centering
    \subfigure[Language Mismatch Ratio]{\includegraphics[width=0.49\linewidth]{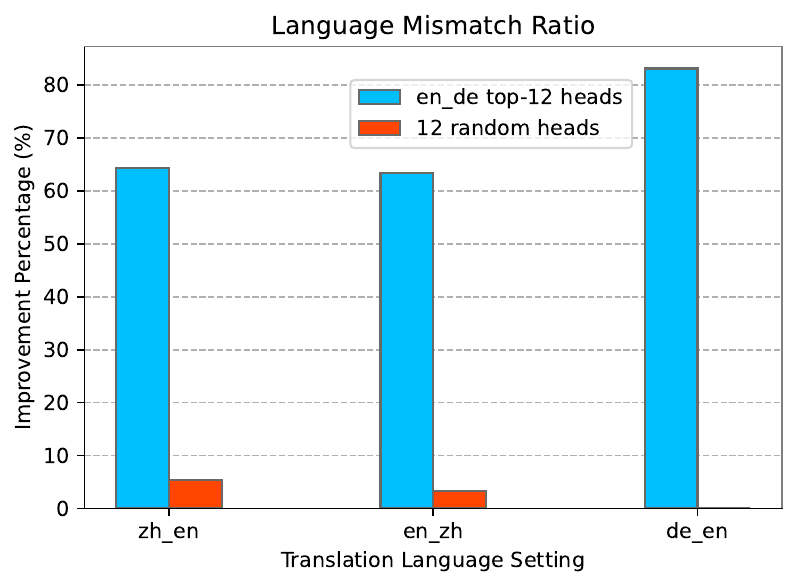}} 
    \subfigure[COMET22DA]{\includegraphics[width=0.49\linewidth]{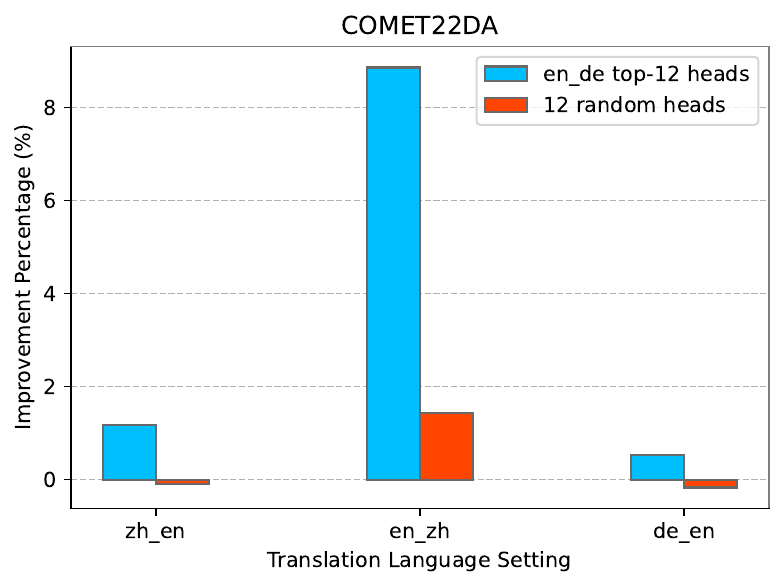}} 
\vspace{-0.1in}
    \caption{Performance ((a) for the decrease percentage of LMR; (b) for the improvement percentage of COMET22DA) of intervention (blue bars) with language settings of \textbf{zh}$\rightarrow$\textbf{en}, \textbf{en}$\rightarrow$\textbf{zh} and \textbf{de}$\rightarrow$\textbf{en} on the heads located with the language setting of  \textbf{en}$\rightarrow$\textbf{de}. The red bars (comparison group) refer to the results for intervention on random heads of the same number.}
    \label{fig:transferring_heads}
\vspace{-0.1in}
\end{figure}
The hypothesis can also be verified to some extent by the important head locating experiments depicted in Figure~\ref{fig:locating_head_main_text}, where results for different language settings (\textbf{de}$\rightarrow$\textbf{en} and \textbf{zh}$\rightarrow$\textbf{en}) share a large proportion of top heads.
Moreover, we locate top-12 important attention heads in LLaMA2-7B under the language setting of en$\rightarrow$de and apply \textit{MT vectors} to LLaMA2-7B with these located heads under the language settings of \textbf{zh}$\rightarrow$\textbf{en}, \textbf{en}$\rightarrow$\textbf{zh} and \textbf{de}$\rightarrow$\textbf{en}.
The results of Zero-Shot translation are depicted in Figure~\ref{fig:transferring_heads} (experimental group, blue bars). 
We additionally randomly select 12 heads to apply \textit{MT vectors} and the results (comparison group) are shown with red bars. 
We observe that for both the language mismatch ratio and COMET22DA\footnote{https://huggingface.co/Unbabel/wmt22-comet-da}, the performance of experimental group largely exceeds the performance of comparison group under all three other language settings, indicating that the attention heads located under a single language setting can transfer to other language settings.
Given these evidences, we propose our first modification to both \textit{MT vectors} and \textit{RP neurons}: \textbf{\textit{We firstly locate attention heads or FFN neurons separately for each language setting and then get the final located results by intersecting the located results for all of language settings.}}
\begin{table}[t]
\vspace{-0.1in}
\centering
\resizebox{\linewidth}{!}{
\begin{tabular}{lccc}
\hline
\textit{\textbf{Zero-Shot}} & \textbf{L}($\downarrow$)  & \textbf{B}($\uparrow$) & \textbf{C}($\uparrow$)  \\
\hline                     
LLaMA2-7B  &$0.0486$&$17.1288$&$0.722$\\
+\textit{MTV}&$\mathbf{-92.46}\%$&$-0.81\%$&$2.65\%$ \\             
+\textit{MTV-I}&$-80.15\%$&$53.5\%$&$15.51\%$ \\
+\textit{MTV-I-D} & $-86.12\%$ & $\mathbf{76.82}\%$  & $\mathbf{16.02}\%$\\
\hline
\hline
\textit{\textbf{One-Shot}} & \textbf{R}($\downarrow$)  & \textbf{B}($\uparrow$) & \textbf{C}($\uparrow$)  \\
\hline                      
LLaMA2-7B  &$0.0035$&$18.8714$&$0.7376$\\  
+\textit{RPN} &$-8.57\%$&$0.07\%$&$\mathbf{0.0}\%$\\
+\textit{RPN-I}&$\mathbf{-25.71}\%$ & $\mathbf{0.51}\%$& $-0.04\%$ \\
\hline
\end{tabular}
}
\vspace{-0.1in}
\caption{
Performance of LLaMA2-7B (and after applying model editing methods) on $\mathcal{D}_{test}$ (under the language settings of zh$\rightarrow$en for \textit{\textbf{Zero-Shot}} and zh$\rightarrow$en for \textit{\textbf{One-Shot}}). Other notations and abbreviations are following Table~\ref{tab:direct_adaptation_main_text}.
}
\vspace{-0.1in}
\label{tab:intersection_main_text}
\end{table}
\begin{table*}[t]
\vspace{-0.1in}
\centering
\resizebox{\textwidth}{!}{
\begin{tabular}{lcccccccc}
\toprule
\textbf{\textit{}}&  \multicolumn{2}{c}{\textbf{de$\rightarrow$en}} & \multicolumn{2}{c}{\textbf{en$\rightarrow$de}} & \multicolumn{2}{c}{\textbf{zh$\rightarrow$en}} & \multicolumn{2}{c}{\textbf{en$\rightarrow$zh}}  \\
\midrule
\midrule
\textbf{\textit{Zero-Shot}} & \textbf{L}($\downarrow$)  & \textbf{B}($\uparrow$)  & \textbf{L}($\downarrow$)  & \textbf{B}($\uparrow$) & \textbf{L}($\downarrow$)  & \textbf{B}($\uparrow$) & \textbf{L}($\downarrow$)  & \textbf{B}($\uparrow$)\\
\midrule                     
LLaMA2-7B   &$0.0219$   & $35.3448$ &$0.4524$  &$12.6084$ &$0.0486$  &$17.1288$ &$0.3269$  &$16.3441$\\
+5-Shot ICL &$-74.89\%$ & $\textbf{4.93\%}$  &$-92.06\%$  &$101.27\%$  &$-50.0\%$  &$\textbf{12.46\%}$  &$\textbf{-82.59\%}$  &$76.9\%$\\
+LoRA       &$\textbf{-83.56\%}$ & $0.68\%$  &$\textbf{-95.25\%}$  &$\textbf{115.24\%}$ &$\textbf{-79.22\%}$  &$6.62\%$  &$-77.58\%$  &$\textbf{82.62\%}$\\
+Full-FT    &$\underline{-8.68\%}$  & $2.25\%$  &$\underline{-62.69\%}$  &$\underline{55.41\%}$   &$\underline{-33.33\%}$  &$3.15\%$  &$\underline{-66.23\%}$  &$62.64\%$ \\
+\textit{MTV-I-D} & $-33.33\%$ & $\underline{-0.53\%}$  & $-86.12\%$& $76.82\%$ & $-54.12\%$& $\underline{-14.08\%}$ & $-69.9\%$& $\underline{24.64\%}$\\
\midrule
\midrule

\textbf{\textit{One-Shot}} & \textbf{R}($\downarrow$)  & \textbf{B}($\uparrow$)  & \textbf{R}($\downarrow$)  & \textbf{B}($\uparrow$) & \textbf{R}($\downarrow$)  & \textbf{B}($\uparrow$) & \textbf{R}($\downarrow$)  & \textbf{B}($\uparrow$)\\
\midrule                    
LLaMA2-7B   &$0.0018$ &$36.5445$ &$0.0141$ &$24.9685$  &$0.0035$ &$18.8714$ &$0.0146$ &$27.7798$\\  
+5-Shot ICL &$0.0\%$    &$\textbf{1.49\%}$    &$\underline{14.89\%}$  &$1.63\%$     &$-14.29\%$ &$2.07\%$ &$-17.12\%$ &$4.08\%$\\
+LoRA       &$\textbf{-77.78\%}$ &$\underline{-9.47\%}$   &$\textbf{-74.47\%}$ &$\underline{-2.39\%}$    &$\underline{5.71\%}$ &$\underline{0.07\%}$ &$-10.27\%$ &$0.37\%$\\
+Full-FT    &$\underline{22.22\%}$  &$1.26\%$    &$-25.53\%$ &$\textbf{4.9\%}$      &$-22.86\%$ &$\textbf{2.5\%}$ &$\underline{\-22.6\%}$ &$\textbf{4.47\%}$\\
+\textit{RPN-I}&$-38.89\%$ & $0.74\%$ &$-27.66\%$ & $0.35\%$& $\textbf{-25.71\%}$& $0.51\%$& $\textbf{-19.18\%}$& $\underline{-0.23\%}$ \\
\bottomrule
\end{tabular}
}
\vspace{-0.1in}
\caption{
Overall Performance of LLaMA2-7B (and after applying model editing methods) on $\mathcal{D}_{test}$ under all language settings. Other notations and abbreviations are following Table~\ref{tab:direct_adaptation_main_text}. The \textbf{bold} value means the best performance while the \underline{underline} value represents the worst performance.
}
\vspace{-0.1in}
\label{tab:overall_main_text}
\end{table*}
We denote the MT vectors fetched by intersected attention heads as \textit{\textbf{MT} \textbf{V}ectors-\textbf{I}ntersection} (\textit{MTV-I}) and intersected RP neurons as \textit{\textbf{R}e\textbf{P}etition \textbf{N}eurons-\textbf{I}ntersection} (\textit{RPN-I}).
We post the results for leveraging \textit{MTV-I} and \textit{RPN-I} under the language settings of en$\rightarrow$de and zh$\rightarrow$en in Table~\ref{tab:intersection_main_text}.
We observe that:
(1) for \textit{MTV-I}, the decrease percentage of language mismatch error ratio ($-80.15\%$) is slightly lower than \textit{MTV} ($-92.46\%$) while improvement percentage of the BLEU score ($53.5\%$) and COMET22DA score ($15.51\%$) exceed \textit{MTV} ($-0.81\%$ and $2.65\%$) by a large margin and
(2) for \textit{RPN-I}, the decrease percentage of repetition error ratio ($-25.71\%$) is much higher than RPN ($-8.57\%$),
suggesting that intersection of different language settings can filter attention heads and FFN neurons that are irrelevant to language mismatch errors and repetition errors out.
On the basis of \textit{MTV-I}, we propose another slight modification: \textbf{\textit{Firstly calculate the MTV-I, then divide it evenly according to the number of the intersected attention heads and add them to those heads}}.
We denote this manner of leveraging \textit{MTV-I} as \textit{MTV-I-Distributional} (\textit{MTV-I-D}). 
We also post the results of leveraging \textit{MTV-I-D} in Table~\ref{tab:intersection_main_text}, where the results demonstrate that \textit{MTV-I-D} can further achiever better performance than \textit{MTV-I} in terms of language mismatch ratio, BLEU and COMET22DA.
\subsection{Overall Results}
\label{sec:modification_results}
To make readers get a better sense of the LLMs edited with our methods (\textit{MTV-I-D} and \textit{RPN-I}), we show the overall evaluation results for both our methods and traditional adaptation methods, including 5-Shot In-Context Learning~ \cite{gpt3} (5-Shot ICL), Low Rank Adaptation Tuning~ \cite{lora} (LoRA) and Full parameter Supervised Fine-Tuning~ \cite{few-shottuningllm-mt} (Full-FT) for LLM-based MT in Table~\ref{tab:overall_main_text}.\footnote{We train the model under Zero-Shot and One-shot respectively except for Five-Shot ICL, other details can be seen on Appendix~\ref{app_implement_details}.}
We post the performance on the metrics of language mismatch error ratio, repetition error ratio and BLEU score (We find that performance on COMET score is highly aligned with BLEU score) and observe that:
(1) Applying the modified editing methods, \textit{MTV-I-D} and \textit{RPN-I} can generally reduce the error ratios for both language mismatch (\textbf{L}) and repetition (\textbf{R}) to a large degree,
(2) The negative side-effect on the general translation quality (BLEU score, \textbf{B}) is minor (except when applying \textit{MTV-I-D} under the setting of \textbf{zh$\rightarrow$en}, with a $-14.08\%$ decrease percentage on BLEU score). It is noteworthy that applying \textit{MTV-I-D} can even improve the general translation quality to a large extent on the settings of \textbf{en$\rightarrow$de} ($76.82\%$) and \textbf{en$\rightarrow$zh} ($24.64\%$) and
(3) The performance of \textit{MTV-I-D} and \textit{RPN-I} can sometimes be comparable with (and even surpass) the traditional methods that adapt LLMs to the MT without additional requirements like long-context prompting and fine-tuning.

Besides, we also investigate \textit{whether applying our editing methods to correct the language mismatch and repetition errors will bring negative effects on the general abilities of LLMs?} and \textit{the additional computation cost brought by applying these editing methods}. The detailed empirical results and analysis are presented in Appendix~\ref{other_discussion}. For the first point, we demonstrate that our editing methods would hardly hurt (sometimes even boost) the general performance of LLMs on five popular benchmarks (MMLU~\cite{benchmarkmmlu}, TruthfulQA~ \cite{benchmarktruthfulqa}, MMLU-Pro~ \cite{benchmarkmmlupro}, CMMLU~ \cite{benchmarkcmmlu}and CommonQA~ \cite{benchmarkcommonsense}). For the second point, through both running time complexity analysis and empirical statistics, we show that the additional computation overhead brought by our editing methods is marginal. We release our code and data on GitHub\footnote{\url{https://github.com/weichuanW/llm-based-mt-via-model-editing}} for the reproduction and exploration of others.

\section{Conclusion}
In the work, we find that two types of errors, language mismatch and repetition, occur frequently when performing machine translation with LLMs, bringing severe negative effects on the translation quality. We investigate the potentials of leveraging model editing methods to mitigate these issues and find that directly adopting function vectors and knowledge neurons may either have limited improvement on the identified errors or bring noteworthy negative effect on the general machine translation quality (e.g., BLEU score), which indicates that the located attention heads and FFN neurons might be too coarse to only affect the error ratios without hurting the translation quality. To this end, we propose to refine the located attention heads and neurons by fetching the intersection of the locating results under different language settings. Our empirical results suggest that the modified function vectors and knowledge neurons methods (\textit{MTV-I-D} and \textit{RPN-I}) can effectively reduce the aforementioned two types of errors and generally bring a positive influence evaluated with the translation quality metrics in most settings, indicating that there indeed exist a small set of modules that are highly responsible for the language mismatch and the repetition errors.
\section*{Limitations}

Our work is based on open-source LLaMA series models LLaMA2\footnote{\url{https://llama.meta.com/llama2/}}. However, the effectiveness of these findings on other models, such as other state-of-the-art open-sourced LLMs (e.g., Mistral-7B\footnote{\url{https://mistral.ai/news/announcing-mistral-7b/}}, OLMO\footnote{\url{https://allenai.org/olmo}} and so on) or the close-sourced LLMs (e.g., GPT-4\footnote{\url{https://openai.com/index/gpt-4/}} and Claude-3.5-Sonnet\footnote{\url{https://www.anthropic.com/news/claude-3-5-sonnet}}), remains under explored. We leave it for the future work.

The model editing methods used in this paper require computational resources proportional to the size of the LLM. When applying our methods to a larger model, more computational resources will be necessary to achieve improved results.
Our focus is on high-resource language settings for machine translation. However, the observations and conclusions may differ when applied to low-resource or non-English language pair settings (e.g., zh→de machine translation)

We utilise automatic metrics for error and machine translation evaluation in our measurements. However, employing human-involved evaluations~\cite{wmt23eval} can offer a more profound understanding of machine translation with LLMs.

\section*{Ethics Statement}
This paper utilizes a pre-trained LLM, with its training data sourced from web corpora that have not undergone ethical filtering. Consequently, it is capable of generating toxic content in the machine translation~\cite{wen-etal-2023-unveiling}. Moreover, we do not filter the source data or translation output in our work. Future research may build on our results to enhance the model, and we advocate for incorporating content supervision to prevent the dissemination of toxic content.

\section*{Acknowledgements}
We sincerely thank the anonymous reviewers for their insightful suggestions to this work. The work was supported by grants from the National Key R\&D Program of China (No.2021ZD0111801), the Start-up Grant (No. 9610564), the Donations for Research Projects (No. 9229129) of the City University of Hong Kong, and the Early Career Scheme (No. CityU 21219323) of the University Grants Committee (UGC), the Research Grants Council of the Hong Kong SAR under Grant GRF 11217823 and Collaborative Research Fund C1042-23GF, the National Natural Science Foundation of China under Grant 62371411, InnoHK initiative, The Government of the HKSAR, and Laboratory for AI-Powered Financial Technologies. 
\bibliography{acl_latex}
\appendix
\section{Dataset Information}\label{app:details_data_construct}
All data used in this work are human-checked from the WMT conference to ensure the data quality. Table ~\ref{tab:dataset_size} shows the detailed data size for $\mathcal{D}_{exps}$, $\mathcal{D}_{train}$ and $\mathcal{D}_{test}$. We use the WMT21 test set\footnote{\url{https://github.com/wmt-conference/wmt21-news-systems}} as the $\mathcal{D}_{exps}$, WMT22 test set\footnote{\url{https://github.com/wmt-conference/wmt22-news-systems}} as $\mathcal{D}_{train}$ and WMT23 test set\footnote{\url{https://github.com/wmt-conference/wmt23-news-systems}} as$\mathcal{D}_{test}$.
\begin{table}[!ht]
\renewcommand{\arraystretch}{1.0}
\resizebox{\linewidth}{!}{
\begin{tabular}{l|l|l|l}
\hline
Setting & $\mathcal{D}_{exps}$ Size & $\mathcal{D}_{train}$ Size & $\mathcal{D}_{test}$ size\\ \hline
en$\rightarrow$de      & 1002      & 2037        & 557               \\ 
de$\rightarrow$en      & 1000      & 1984        & 549               \\ 
en$\rightarrow$zh      & 1002      & 2037        & 2074              \\ 
zh$\rightarrow$en      & 1948      & 1875        & 1976              \\ \hline
\end{tabular}}
\caption{Data size of $\mathcal{D}_{exps}$, $\mathcal{D}_{train}$, $\mathcal{D}_{test}$ on four language settings.}
\label{tab:dataset_size}
\end{table}

The detailed data size for the $K$-shot ($K=0, 1,5$) setting is shown in Table~\ref{tab:dataset_size_shots}. For all settings, we use the \textit{lang prompt} as the prompt template ( as shown in Section~\ref{sec:data}). For the Zero-Shot setting, we directly combine the source data with the \textit{lang prompt}. For the One-Shot setting, we uniformly sample the data from $\mathcal{D}_{exps}$ based on the length of the example source to alleviate the potential length bias from prompt example \cite{prompt-llmmt}. We use the most natural selection method for the Five-Shot setting by randomly selecting five examples from $\mathcal{D}_{exps}$.
\begin{table}[!ht]
\renewcommand{\arraystretch}{1.0}
\resizebox{\linewidth}{!}{
\begin{tabular}{l|l|l|l}
\hline
Setting & $\mathcal{D}_{0}$ Size & $\mathcal{D}_{1}$ Size & $\mathcal{D}_{5}$ size\\ \hline
en$\rightarrow$de ($\mathcal{D}_{train}$)&2037 & 12222        & 2037               \\ 
de$\rightarrow$en ($\mathcal{D}_{train}$)& 1984 & 9920      & 1984               \\ 
en$\rightarrow$zh ($\mathcal{D}_{train}$) & 2037 & 12222        & 2037              \\ 
zh$\rightarrow$en ($\mathcal{D}_{train}$)& 1875  & 11250        & 1875              \\ \hline
en$\rightarrow$de ($\mathcal{D}_{test}$)     & 557     & 3342        & 557               \\ 
de$\rightarrow$en ($\mathcal{D}_{test}$)     & 549      & 2745      & 549               \\ 
en$\rightarrow$zh ($\mathcal{D}_{test}$)     & 2074      & 12444        & 2074              \\ 
zh$\rightarrow$en ($\mathcal{D}_{test}$)     & 1976      & 11856        & 1976              \\ \hline
\end{tabular}}
\caption{Data size of Zero-Shot ($\mathcal{D}_{0}$), One-Shot($\mathcal{D}_{1}$) and Five-Shot($\mathcal{D}_{5}$) on four language settings. $\mathcal{D}_{train}$ and $\mathcal{D}_{test}$ represent the source data in the prompt.}
\label{tab:dataset_size_shots}
\end{table}

\section{Toolkits for evaluation}\label{app:details_eval}
We use spBLEU~\cite{post-2018-call} and COMET22DA~\cite{rei-etal-2022-comet} from huggingface API\footnote{https://huggingface.co/docs/evaluate/index} to evaluate MT quality.
For the language mismatch detection, we use the Polyglot toolkit\footnote{\url{https://github.com/aboSamoor/polyglot}} to detect the language error.
For repetition error, based on the definition of repetition error, we follow two rules to judge whether a translation result is repeated: (1) the generation length reaches the $max\_new\_tokens$ setting\footnote{\url{https://github.com/huggingface/tokenizers}} (We use 400 in our work); (2) there exists a substring happening until the end of the generation.
For the machine translation metrics, we use SacreBLEU \cite{post-2018-call}, Unbabel/wmt22-comet-da\footnote{\url{https://huggingface.co/Unbabel/wmt22-comet-da}} and Unbabel/wmt22-cometkiwi-da\footnote{\url{https://huggingface.co/Unbabel/wmt22-cometkiwi-da}} to do evaluation.

\section{Comprehensive exploring of language mismatch and repetition errors}\label{app_compre_prompts}
When leveraging LLM for translation, prompts matter a lot. To comprehensively understand the effect of prompt templates, we include multiple templates and with the following formats:
\begin{table}[!h]
\vspace{-0.1in}
\centering
\resizebox{0.47\textwidth}{!}{
\begin{tabular}{l|l}
\hline
\textbf{Template} & \textbf{Prompt}                                          \\ \hline\hline
Ours     & L1: $src_q$ \textbackslash{}n L2: \\ \hline
Temp1 & Given the following source text: $src_q$, a good L2 translation is:  \\ \hline
Temp2 & If the original version says $src_q$ then the L2 version should say: \\ \hline
Temp3 & What is the L2 translation of the sentence: $src_q$?                 \\ \hline
Temp4    & L1: $src_q$ = L2:                 \\ \hline
Temp5    & $src_q$ translates into L2 as:    \\ \hline
\end{tabular}}
\caption{Different template formats for Zero-Shot Settings. \textit{L1} and \textit{L2} represent for the source language and target language respectively. $src_q$ means the source sentence to be translated.}
\label{prompt_templates}
\end{table}

In Table~\ref{prompt_templates}, we include six templates, including the template used in this paper (Ours~ \cite{palmmt}). And other five templates from PaLM \cite{PaLMmodel}, PaLM \cite{PaLMmodel}, GPT-3 \cite{gpt3}, XGLM \cite{xglm} and CoT prompting \cite{CoTprompt} respectively. For the One-Shot settings, we use a similar way of \textit{Lang Prompt} in Section~\ref{sec:preliminary} by adding exemplars to the same template as the One-Shot prompt. We further evaluate the Language Mismatch Ratio (Zero-Shot) and Repetition Ratio (One-Shot) of LLaMA2-7B \cite{llama2} under the en$\rightarrow$zh setting.
Table~\ref{LR_RR_templates_enzh} demonstrates that the identified errors are general even if we change the prompt format.
\begin{table}[!h]
\centering
\begin{tabular}{l|l|l}
\hline
\textbf{Template} & \textbf{L$\downarrow$} & \textbf{R$\downarrow$}       \\ \hline\hline
Ours     & 0.3269 & 0.0146 \\ \hline
Temp1    & 0.2864 & 0.0133 \\ \hline
Temp2    & 0.5598 & 0.0134 \\ \hline
Temp3    & 0.9904 & 0.0161 \\ \hline
Temp4    & 0.3896 & 0.0138 \\ \hline
Temp5    & 0.6962 & 0.0127 \\ \hline
\end{tabular}
\caption{The language mismatch rato and repetition ratio on six different templates of en$\rightarrow$zh under Zero-Shot Settings and One-shot Settings, respectively. \textbf{L}: language mismatch
ratio; \textbf{R}: repetition ratio.}
\label{LR_RR_templates_enzh}
\end{table}

In this paper. We demonstrate that our detected MT vector and MT neurons are effective for our default setting across different language settings. Another interesting question is: \textbf{\textit{Are our detected MT vector, MT neurons and our proposed methods general if we change the prompts?}}

Table~\ref{improve_other_prompts_enzh} can answer this question to some extent. The first surprising thing is that they are generally effective despite prompt shifts. For example, \textbf{Temp3} has a nearly total failure in language mismatch (almost 100\% mismatch), while our methods can still improve this error with 8.56\%. Besides, our detected repetition neurons are pretty helpful for the repetition error across different prompt templates. Another surprising thing is that although we use our prompt setting directly to extract the MT vector or repetition neurons, it is effective on all prompts. This indicates that these detected key components are highly related to MT rather than prompts. A further study on the intersection of MT vectors and repetition neurons among different template prompts will be one of our next steps.

\begin{table}[!h]
\centering
\resizebox{0.49\textwidth}{!}{
\begin{tabular}{l|l|l|l|l}
\hline
\textbf{Template} & \textbf{L$\downarrow$} & \textbf{MTV-I-D}  & \textbf{R$\downarrow$}&  \textbf{RPN-I}       \\ \hline
Temp1    & 0.2864 & -79.80\% & 0.0133 & -49.40\% \\ \hline
Temp2    & 0.5598 & -72.18\% & 0.0134 & -56.29\% \\ \hline
Temp3    & 0.9904 & -8.56\%  & 0.0161 & -47.00\% \\ \hline
Temp4    & 0.3896 & -53.47\% & 0.0138 & -55.81\% \\ \hline
Temp5    & 0.6962 & -60.66\% & 0.0127 & -60.75\% \\ \hline
\end{tabular}}
\caption{The effect of our proposed methods on other prompts on en$\rightarrow$zh setting. Other notations and abbreviations are following Table~\ref{tab:direct_adaptation_main_text}.}
\label{improve_other_prompts_enzh}
\end{table}

\section{The AIE values for all heads}\label{app:all_aies}
\begin{figure}[!ht]
    \centering
    \subfigure[en$\rightarrow$de]{\includegraphics[width=0.49\linewidth]{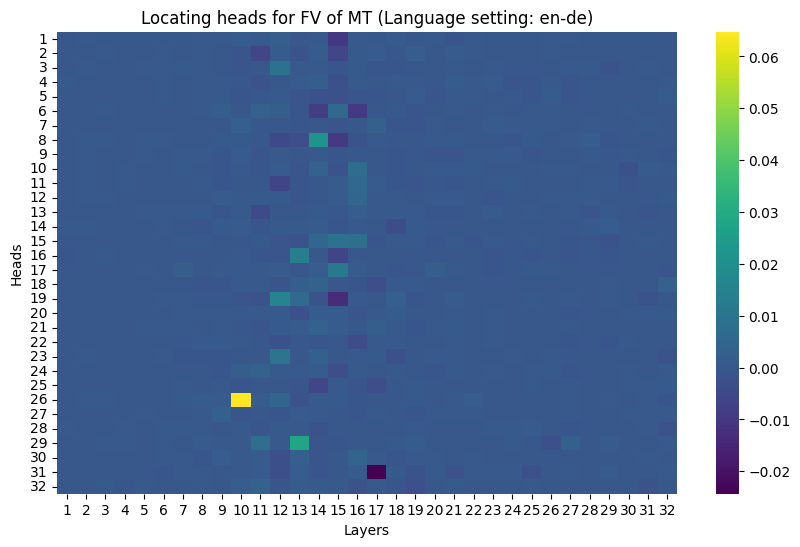}} 
    \subfigure[de$\rightarrow$en]{\includegraphics[width=0.49\linewidth]{figures/all_heads_aie_lang/all_fv_heatmap_de-en_1024.png}} 
    \subfigure[en$\rightarrow$zh]{\includegraphics[width=0.49\linewidth]{figures/all_heads_aie_lang/all_fv_heatmap_en-zh_1024.png}} 
    \subfigure[zh$\rightarrow$en]{\includegraphics[width=0.49\linewidth]{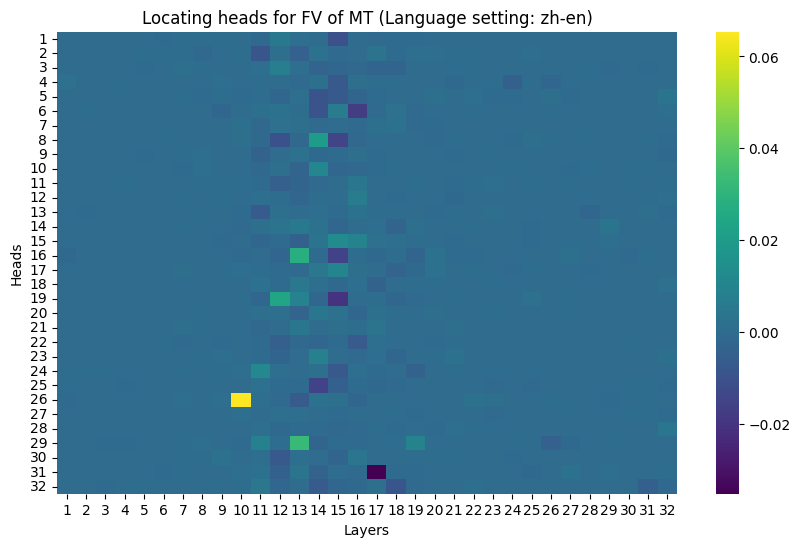}} 
\vspace{-0.1in}
    \caption{Heatmaps of AIE values for attention heads in LLaMA2-7B for en$\rightarrow$de setting (a), de$\rightarrow$en setting (b), en$\rightarrow$zh setting (c) and zh$\rightarrow$en setting (d). The x-axis and y-axis refer to the layer and head, respectively. Brighter color refers to the head with larger AIE value.}
    \label{fig:all_aies_langs}
\vspace{-0.1in}
\end{figure}

\begin{figure}[!ht]
    \centering
    \subfigure[en$\rightarrow$de]{\includegraphics[width=0.49\linewidth]{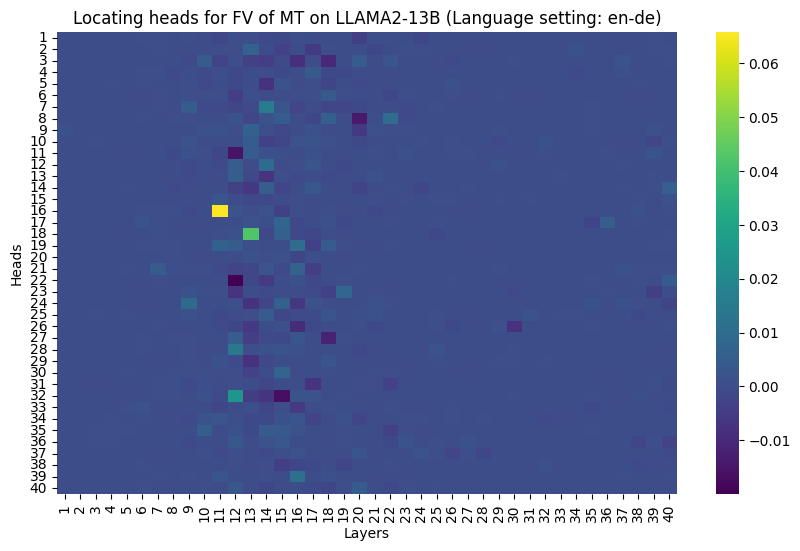}} 
    \subfigure[de$\rightarrow$en]{\includegraphics[width=0.49\linewidth]{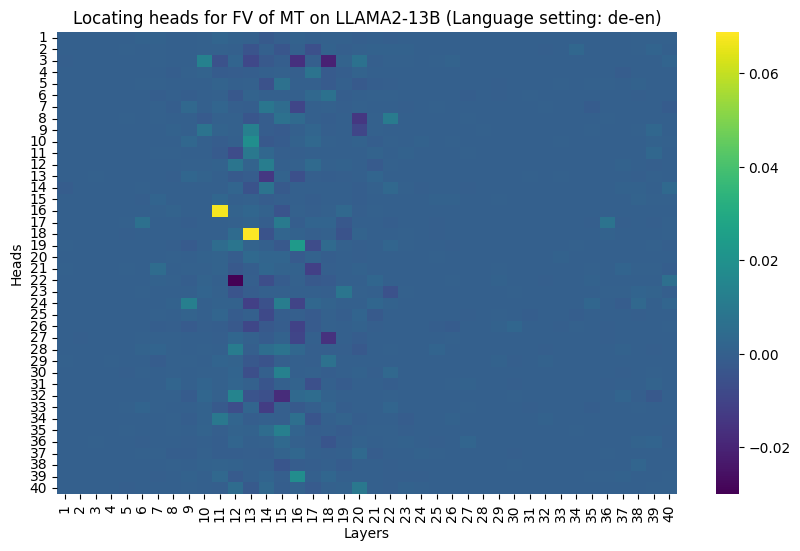}} 
    \subfigure[en$\rightarrow$zh]{\includegraphics[width=0.49\linewidth]{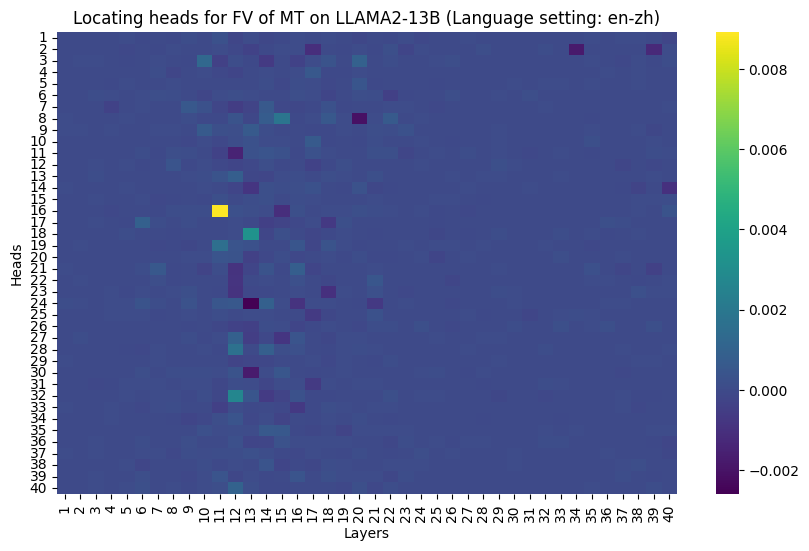}} 
    \subfigure[zh$\rightarrow$en]{\includegraphics[width=0.49\linewidth]{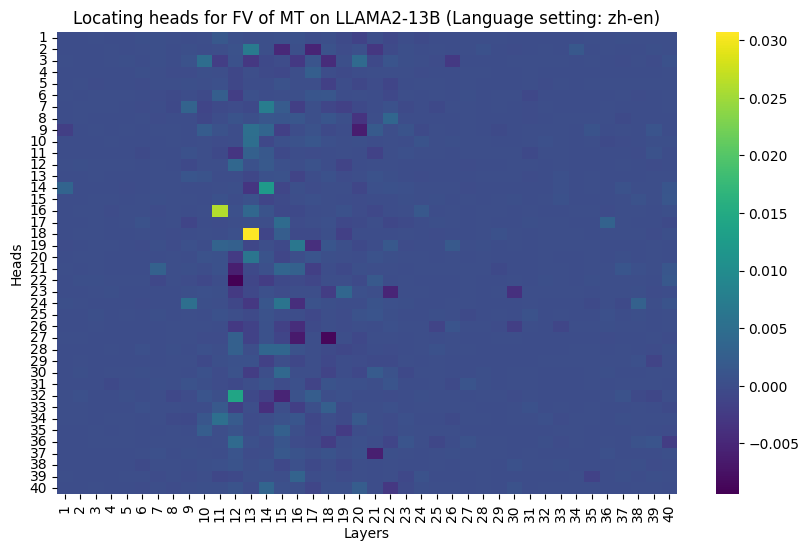}} 
\vspace{-0.1in}
    \caption{Heatmaps of AIE values for attention heads in LLaMA2-13B for en$\rightarrow$de setting (a), de$\rightarrow$en setting (b), en$\rightarrow$zh setting (c) and zh$\rightarrow$en setting (d). The x-axis and y-axis refer to the layer and head, respectively. Brighter color refers to the head with larger AIE value.}
    \label{fig:all_aies_langs_13b}
\vspace{-0.1in}
\end{figure}

For a comprehensive observation and validation of our independent assumption for machine translation heads. We show all language setting results in Figure~\ref{fig:all_aies_langs}. The AIE values of all heads of LLaMA2-7B include en$\rightarrow$de, de$\rightarrow$en, en$\rightarrow$zh and zh$\rightarrow$en settings. We also include LLaMA2-13B, a larger LLM than LLaMA2-7B in the same LLM family. Figure~\ref{fig:all_aies_langs_13b} shows the AIE values of all heads of LLaMA2-13B. 

We can observe sets of overlapped heads on both LLaMA2-7B and LLaMA2-13B. This indicates that our detected MT heads may be a crucial mechanism for LLMs when processing MT. Another valuable information from these figures is the locations of these heads are on the middle layers, which can also be found and explored by other current multilingual research \cite{multilingual1, multilingual2}. Further analysis and concrete results are shown in the Appendix~\ref{scaling_exps}.

Figure \ref{fig:all_aies_langs_13b} shows the AIE values of all heads of LLaMA2-13B on en$\rightarrow$de, de$\rightarrow$en, en$\rightarrow$zh and zh$\rightarrow$en settings.

\section{Results for direct adaptation}\label{app_direct_adapt}
The complete results of direct adaptation on four language settings are shown in Table \ref{tab:direct_adaptation_en_de} (en$\rightarrow$de), \ref{tab:direct_adaptation_de_en} (de$\rightarrow$en), \ref{tab:direct_adaptation_en_zh} (en$\rightarrow$zh) and \ref{tab:direct_adaptation_zh_en} (zh$\rightarrow$de). 

These tables show that the MT vectors can decrease the language mismatch ratio while the RP neurons help decrease repetition errors in all language settings. One notable case is on Table~\ref{tab:direct_adaptation_de_en}, where we do not extract the \textit{RP neurons} since there is no repetition error on the $\mathcal{D}_{train}$. However, we do observe repetition errors when applying the data sourced from $\mathcal{D}_{test}$. This phenomenon indicates that sometimes, the error is easily ignored when we use out-of-date data or a small set of data to check. Our next step is to improve the robustness of our proposed method by exploring the potential properties of these error prompts on a data view.

\begin{table}[!ht]
\centering
\resizebox{0.49\textwidth}{!}{
\begin{tabular}{lccc}
\hline
\textit{\textbf{Zero-Shot}} & \textbf{L}($\downarrow$)  & \textbf{B}($\uparrow$) & \textbf{C}($\uparrow$)  \\
\hline                     
LLaMA2-7B  &$0.4524$&$12.6084$&$ 0.6113$\\
+\textit{MT vectors} &$-92.46\%$&$-0.81\%$&$2.65\%$\\                
+\textit{MT neurons} &$-11.1\%$&$1.78\%$&$0.15\%$\\
\hline
\hline
\textit{\textbf{One-Shot}} & \textbf{R}($\downarrow$)  & \textbf{B}($\uparrow$) & \textbf{C}($\uparrow$)  \\
\hline                      
LLaMA2-7B  &$0.0141$&$24.9685$&$0.7279$\\  
+\textit{MT vectors} &$487.94\%$&$-39.11\%$&$-10.87\%$\\                
+\textit{MT neurons} &$4.26\%$&$-1.05\%$&$-1.06\%$\\
+\textit{RP neurons} &$-27.66\%$&$0.77\%$&$-0.3\%$\\
\hline
\end{tabular}}
\vspace{-0.1in}
\caption{
Performance of LLaMA2-7B (and after applying model editing methods) on $\mathcal{D}_{test}$ (under the language setting of \textbf{en$\rightarrow$de}). \textit{\textbf{Zero-Shot}} and \textit{\textbf{One-Shot}} refer to using a Zero-Shot prompt (for language mismatch errors) and one-shot prompt (for repetition errors) for MT tasks. For evaluation metrics, \textbf{L}: Language mismatch ratio; \textbf{R}: Repetition ratio; \textbf{B}: BLEU and \textbf{C}: COMET22DA, where \textbf{B} and \textbf{C} mainly evaluate the general translation quality. For plain LLaMA2-7B, the results are absolute values; for LLaMA2-7B with editing methods, the results are relative \textbf{improvement percentages}.
}
\vspace{-0.1in}
\label{tab:direct_adaptation_en_de}
\end{table}

\begin{table}[!ht]
\centering
\resizebox{0.49\textwidth}{!}{
\begin{tabular}{lccc}
\hline
\textit{\textbf{Zero-Shot}} & \textbf{L}($\downarrow$)  & \textbf{B}($\uparrow$) & \textbf{C}($\uparrow$)  \\
\hline                     
LLaMA2-7B  &$0.0219$&$35.3448$&$ 0.7836$\\
+\textit{MT vectors} &$-74.89\%$&$-33.85\%$&$-5.53\%$\\                
+\textit{MT neurons} &$8.22\%$&$0.03\%$&$0.23\%$\\
\hline
\hline
\textit{\textbf{One-Shot}} & \textbf{R}($\downarrow$)  & \textbf{B}($\uparrow$) & \textbf{C}($\uparrow$)  \\
\hline                      
LLaMA2-7B  &$0.0018$&$36.5445$&$0.7893$\\  
+\textit{MT vectors} &$727.78\%$&$-33.62\%$&$-4.38\%$\\                
+\textit{MT neurons} &$22.22\%$&$-0.35\%$&$-0.11\%$\\
+\textit{RP neurons} &$--\%$&$--\%$&$--\%$\\
\hline
\end{tabular}}
\vspace{-0.1in}
\caption{
Performance of LLaMA2-7B (and after applying model editing methods) on $\mathcal{D}_{test}$ (under the language setting of \textbf{de$\rightarrow$en}). The \textbf{$--$} means the same result as the LLaMA2-7B since we do not detect any repetition on the training set under the same language setting. Notation and corresponding explanations can refer to Table~\ref{tab:direct_adaptation_en_de}.}
\vspace{-0.1in}
\label{tab:direct_adaptation_de_en}
\end{table}

\begin{table}[!ht]
\centering
\resizebox{0.49\textwidth}{!}{
\begin{tabular}{lccc}
\hline
\textit{\textbf{Zero-Shot}} & \textbf{L}($\downarrow$)  & \textbf{B}($\uparrow$) & \textbf{C}($\uparrow$)  \\
\hline                     
LLaMA2-7B  &$0.3269$&$16.3441$&$0.6567$\\
+\textit{MT vectors} &$-70.05\%$&$18.2\%$&$5.07\%$\\                
+\textit{MT neurons} &$-5.32\%$&$3.16\%$&$0.35\%$\\
\hline
\hline
\textit{\textbf{One-Shot}} & \textbf{R}($\downarrow$)  & \textbf{B}($\uparrow$) & \textbf{C}($\uparrow$)  \\
\hline                      
LLaMA2-7B  &$0.0146$&$27.7798$&$0.7444$\\  
+\textit{MT vectors} &$162.33\%$&$-15.29\%$&$-4.0\%$\\                
+\textit{MT neurons} &$5.48\%$&$-4.28\%$&$-0.28\%$\\
+\textit{RP neurons} &$-4.11\%$&$0.55\%$&$0.05\%$\\
\hline
\end{tabular}}
\vspace{-0.1in}
\caption{
Performance of LLaMA2-7B (and after applying model editing methods) on $\mathcal{D}_{test}$ (under the language setting of \textbf{en$\rightarrow$zh}). \textit{\textbf{Zero-Shot}} and \textit{\textbf{One-Shot}} refer to using a Zero-Shot prompt (for language mismatch errors) and one-shot prompt (for repetition errors) for MT tasks. Notation and corresponding explanations can refer to Table~\ref{tab:direct_adaptation_en_de}.
}
\vspace{-0.1in}
\label{tab:direct_adaptation_en_zh}
\end{table}

\begin{table}[!ht]
\centering
\resizebox{0.49\textwidth}{!}{
\begin{tabular}{lccc}
\hline
\textit{\textbf{Zero-Shot}} & \textbf{L}($\downarrow$)  & \textbf{B}($\uparrow$) & \textbf{C}($\uparrow$)  \\
\hline                     
LLaMA2-7B  &$0.0486$&$17.1288$&$0.722$\\
+\textit{MT vectors} &$-72.84\%$&$-37.35\%$&$-1.84\%$\\                
+\textit{MT neurons} &$-18.72\%$&$4.28\%$&$-0.15\%$\\
\hline
\hline
\textit{\textbf{One-Shot}} & \textbf{R}($\downarrow$)  & \textbf{B}($\uparrow$) & \textbf{C}($\uparrow$)  \\
\hline                      
LLaMA2-7B  &$0.0035$&$18.8714 $&$0.7376$\\  
+\textit{MT vectors} &$482.86\%$&$-23.07\%$&$-1.68\%$\\                
+\textit{MT neurons} &$0.0\%$&$-0.35\%$&$-0.03\%$\\
+\textit{RP neurons} &$-8.57\%$&$0.07\%$&$0.0\%$\\
\hline
\end{tabular}}
\vspace{-0.1in}
\caption{
Performance of LLaMA2-7B (and after applying model editing methods) on $\mathcal{D}_{test}$ (under the language setting of \textbf{zh$\rightarrow$en}). \textit{\textbf{Zero-Shot}} and \textit{\textbf{One-Shot}} refer to using a Zero-Shot prompt (for language mismatch errors) and one-shot prompt (for repetition errors) for MT tasks. Notation and corresponding explanations can refer to Table~\ref{tab:direct_adaptation_en_de}.
}
\vspace{-0.1in}
\label{tab:direct_adaptation_zh_en}
\end{table}

\section{Results for improved adaptation}
Table \ref{tab:intersection_en_de}, \ref{tab:intersection_de_en}, \ref{tab:intersection_en_zh} and \ref{tab:intersection_zh_en} show the results for improved adaptation on en$\rightarrow$de, de$\rightarrow$en, en$\rightarrow$zh and zh$\rightarrow$en respectively. Our proposed \textit{RPN-I} generally show stable improvements in all language settings. For Table~\ref{tab:intersection_de_en}, we skip the \textbf{RPN} and \textbf{RPN-I} since we do not detect repetition errors with $\mathcal{D}_{train}$. Even though we observe stable improvements across all language settings on the language mismatch when applying our proposed \textit{MTV-I-D}, an inevitable decrease happens on X$\rightarrow$en (X refers to de or zh in our work) on any machine translation heads application. Considering LLaMA2 is an English-centric large language model, we think English may not only be used as language recognition but also for other potential mechanisms like general concepts \cite{multilingual1, multilingual2}. This interesting phenomenon can be connected to multilingual research for further exploration.
   
\begin{table}[!ht]
\centering
\resizebox{0.49\textwidth}{!}{
\begin{tabular}{lccc}
\hline
\textit{\textbf{Zero-Shot}} & \textbf{L}($\downarrow$)  & \textbf{B}($\uparrow$) & \textbf{C}($\uparrow$)  \\
\hline                     
LLaMA2-7B  &$0.0219$&$35.3448$&$0.7836$\\
+\textit{MTV}&$\mathbf{-74.89}\%$&$-33.85\%$&$\mathbf{0.0036} \%$ \\             
+\textit{MTV-I}&$-58.45\%$&$-4.84\%$&$-5.53\%$ \\
+\textit{MTV-I-D} & $-33.33\%$ & $\mathbf{-0.53}\%$  & $-0.22\%$\\
\hline
\hline
\textit{\textbf{One-Shot}} & \textbf{R}($\downarrow$)  & \textbf{B}($\uparrow$) & \textbf{C}($\uparrow$)  \\
\hline                      
LLaMA2-7B  &$0.0018$&$36.5445$&$0.7893$\\  
+\textit{RPN} &$--\%$&$--\%$&$--\%$\\
+\textit{RPN-I}&$\mathbf{--}\%$ & $\mathbf{--}\%$& $--\%$ \\
\hline
\end{tabular}}
\vspace{-0.1in}
\caption{
Performance of LLaMA2-7B (and after applying model editing methods) on $\mathcal{D}_{test}$ (under the language settings of de$\rightarrow$en for \textit{\textbf{Zero-Shot}} and de$\rightarrow$en for \textit{\textbf{One-Shot}}). The \textbf{--} means the results is the same as the LLaMA2-7B since there is no repetition cases in the $\mathcal{D}_{train}$. Other notations and abbreviations following Table~\ref{tab:direct_adaptation_en_de}.
}
\vspace{-0.1in}
\label{tab:intersection_de_en}
\end{table}

       
\begin{table}[!ht]
\centering
\resizebox{0.49\textwidth}{!}{
\begin{tabular}{lccc}
\hline
\textit{\textbf{Zero-Shot}} & \textbf{L}($\downarrow$)  & \textbf{B}($\uparrow$) & \textbf{C}($\uparrow$)  \\
\hline                     
LLaMA2-7B  &$0.4524$&$12.6084$&$0.6113$\\
+\textit{MTV}&$\mathbf{-92.46}\%$&$-0.81\%$&$2.65\%$ \\             
+\textit{MTV-I}&$-80.15\%$&$53.5\%$&$15.51\%$ \\
+\textit{MTV-I-D} & $-86.12\%$ & $\mathbf{76.82}\%$  & $\mathbf{16.02}\%$\\
\hline
\hline
\textit{\textbf{One-Shot}} & \textbf{R}($\downarrow$)  & \textbf{B}($\uparrow$) & \textbf{C}($\uparrow$)  \\
\hline                      
LLaMA2-7B  &$0.0141$&$24.9685$&$0.7279$\\  
+\textit{RPN} &$-27.66\%$&$\mathbf{0.77}\%$&$-0.3\%$\\
+\textit{RPN-I}&$\mathbf{-27.66}\%$ & $0.35\%$& $\mathbf{-0.03}\%$ \\
\hline
\end{tabular}}
\vspace{-0.1in}
\caption{
Performance of LLaMA2-7B (and after applying model editing methods) on $\mathcal{D}_{test}$ (under the language settings of en$\rightarrow$de for \textit{\textbf{Zero-Shot}} and en$\rightarrow$de for \textit{\textbf{One-Shot}}). Other notations and abbreviations following Table~\ref{tab:direct_adaptation_en_de}.
}
\vspace{-0.1in}
\label{tab:intersection_en_de}
\end{table}

\begin{table}[!ht]
\centering
\resizebox{0.49\textwidth}{!}{
\begin{tabular}{lccc}
\hline
\textit{\textbf{Zero-Shot}} & \textbf{L}($\downarrow$)  & \textbf{B}($\uparrow$) & \textbf{C}($\uparrow$)  \\
\hline                     
LLaMA2-7B  &$0.0486$&$17.1288$&$0.722$\\
+\textit{MTV}&$\mathbf{-72.84}\%$&$-37.35\%$&$-1.84\%$ \\             
+\textit{MTV-I}&$-54.12\%$&$-20.75\%$&$0.0\%$ \\
+\textit{MTV-I-D} & $-54.12\%$ & $\mathbf{-14.08}\%$  & $\mathbf{0.36}\%$\\
\hline
\hline
\textit{\textbf{One-Shot}} & \textbf{R}($\downarrow$)  & \textbf{B}($\uparrow$) & \textbf{C}($\uparrow$)  \\
\hline                      
LLaMA2-7B  &$0.0035$&$18.8714$&$0.7376$\\  
+\textit{RPN} &$-8.57\%$&$0.07\%$&$\mathbf{0.0}\%$\\
+\textit{RPN-I}&$\mathbf{-25.71}\%$ & $\mathbf{0.51}\%$& $-0.04\%$ \\
\hline
\end{tabular}}
\vspace{-0.1in}
\caption{
Performance of LLaMA2-7B (and after applying model editing methods) on $\mathcal{D}_{test}$ (under the language settings of zh$\rightarrow$en for \textit{\textbf{Zero-Shot}} and zh$\rightarrow$en for \textit{\textbf{One-Shot}}). Other notations and abbreviations are following Table~\ref{tab:direct_adaptation_zh_en}.
}
\vspace{-0.1in}
\label{tab:intersection_zh_en}
\end{table}

\begin{table}[!ht]
\centering
\resizebox{0.49\textwidth}{!}{
\begin{tabular}{lccc}
\hline
\textit{\textbf{Zero-Shot}} & \textbf{L}($\downarrow$)  & \textbf{B}($\uparrow$) & \textbf{C}($\uparrow$)  \\
\hline                     
LLaMA2-7B  &$0.3269$&$16.3441$&$0.6567$\\
+\textit{MTV}&$\mathbf{-70.05}\%$&$18.2\%$&$5.07\%$ \\             
+\textit{MTV-I}&$ -67.27\%$&$19.08\%$&$7.54\%$ \\
+\textit{MTV-I-D} & $-69.9\%$ & $\mathbf{24.64}\%$  & $\mathbf{8.82}\%$\\
\hline
\hline
\textit{\textbf{One-Shot}} & \textbf{R}($\downarrow$)  & \textbf{B}($\uparrow$) & \textbf{C}($\uparrow$)  \\
\hline                      
LLaMA2-7B  &$0.0146$&$27.7798$&$0.7444$\\  
+\textit{RPN} &$-4.11\%$&$0.55\%$&$\mathbf{0.05}\%$\\
+\textit{RPN-I}&$\mathbf{-19.18}\%$ & $0.01\%$& $-0.23\%$ \\
\hline
\end{tabular}}
\vspace{-0.1in}
\caption{
Performance of LLaMA2-7B (and after applying model editing methods) on $\mathcal{D}_{test}$ (under the language settings of en$\rightarrow$zh for \textit{\textbf{Zero-Shot}} and en$\rightarrow$zh for \textit{\textbf{One-Shot}}). Other notations and abbreviations are following Table~\ref{tab:direct_adaptation_en_de}.
}
\vspace{-0.1in}
\label{tab:intersection_en_zh}
\end{table}

\begin{table*}[!h]
\vspace{-0.1in}
\centering
\resizebox{\textwidth}{!}{
\begin{tabular}{lcccccccc}
\toprule
\textbf{\textit{}}&  \multicolumn{2}{c}{\textbf{de$\rightarrow$en}} & \multicolumn{2}{c}{\textbf{en$\rightarrow$de}} & \multicolumn{2}{c}{\textbf{zh$\rightarrow$en}} & \multicolumn{2}{c}{\textbf{en$\rightarrow$zh}}  \\
\midrule
\midrule
\textbf{\textit{One-Shot}} & \textbf{R}($\downarrow$)  & \textbf{B}($\uparrow$)  & \textbf{R}($\downarrow$)  & \textbf{B}($\uparrow$) & \textbf{R}($\downarrow$)  & \textbf{B}($\uparrow$) & \textbf{R}($\downarrow$)  & \textbf{B}($\uparrow$)\\

\midrule                     
LLaMA2-7B  &$1.0$   & $6.1$ &$1.0$  &$12.64$ &$1.0$  &$2.13$ &$1.0$  &$2.08$\\
\midrule 
+case-RM  &$0.0\%$ & $-46.51\%$ &$-35.29\%$ & $-18.25\%$  &$-28.57\%$ &$27.09\%$ &$-31.79\%$ & $43.7\%$\\
+case-FM  &$-33.33\%$ & $-39.75\%$  &$-64.71\%$ & $-4.92\%$&$-60.71\%$ & $148.54\%$ &$-64.16\%$ & $181.75\%$\\
\bottomrule
\end{tabular}
}
\vspace{-0.1in}
\caption{
Case-editing on Repetition cases. The \textit{case-RM} means we detect the RPN on the first repetition region and try to do modifications to remove the first repetition region. The \textit{case-FM} means we detect the repetition region on the first repetition token only and do the modification.
}
\vspace{-0.1in}
\label{tab:case_editing}
\end{table*}

\section{Implementation Details}\label{app_implement_details}
For all machine translation results on LLMs, we recognise the end of the generation through the \textit{line break} based on the format design of \textit{lang prompt}. In the real translation process, we use batch generation techniques (batch size = 4) and set the maximum generation length of tokens to 400 with the Huggingface API\footnote{\url{https://huggingface.co/}} to do translations for all settings in this work. 

\paragraph{Five-Shot} For the Five-Shot setting, we directly use the $\mathcal{D}_{5}$ on LLaMA2-7B to run machine translation task without intervention.

\paragraph{LoRA fine-tuning} LoRA (Low-Rank Adaptation) \cite{lora} is a parameter-efficient tuning technique generally used in natural language processing. In our work, we use the LoRA \cite{lora} method to align the LLaMA2-7B model to the machine translation task. For the fine-tuning data, we combine the data of all language settings from $\mathcal{D}_{train}$ into $\mathcal{D}_{0}$ and $\mathcal{D}_{1}$ for Zero-Shot setting and One-Shot setting respectively (this means we train two different models for Zero-Shot and One-Shot with the same source data from  $\mathcal{D}_{train}$). Finally, we tune two LoRA models with the trl tool\footnote{\url{https://github.com/huggingface/trl}} with the self-supervised tuning method combined with \textit{lang prompt}. We train one epoch with a rank of 64 and a learning rate of $2e^{-4}$ for both Zero-Shot and One-Shot. We use one NVIDIA A100 80GB Tensor Core GPU card for the SFT training; either the Zero-Shot or One-Shot costs less than a half day.
\paragraph{Full fine-tuning} We use the same data and training tool in the LoRA setting for full fine-tuning. In the training process, we use the bfloat16 precious to train the model on one NVIDIA A100 80GB Tensor Core GPU card for full fine-tuning with a lower learning rate $1e^{-6}$ compared to LoRA. 

We claim that there is still room for improvements in the LoRA or Full fine-tuning methods. However, a complete understanding of the mismatch and repetition error should also be evaluated on large-scale data, which is one of the following steps for our research.

\paragraph{The effect on inherent abilities of LLMs.} We use five-shot prompts for all benchmarks except for TruthfulQA \cite{benchmarktruthfulqa} where we adopt the zero-shot prompts. All the experiments are implemented based on our codes and the open-sourced LLMs evaluation harness repository.\footnote{https://github.com/EleutherAI/lm-evaluation-harness/tree/main}

\paragraph{Detected MT heads and Repetition Neurons.} To facilitate the following research on LLM-based MT, we also provide the detailed MT heads and Repetition Neurons detected in this work. For LLaMA2-7B, we find the following overlapped heads after doing an intersection on top-100 AIE heads of each language setting: [[9, 25], [12, 28], [13, 7], [11, 18], [12, 15], [14, 14], [11, 2], [15, 10], [14, 5], [10, 31], [12, 20], [16, 1]], where the first coordinate represents the head index $\in [0,31]$ and the second coordinate represents the layer $\in [0,31]$. For the Repetition Neurons, after choosing the top 300 repetition neurons and doing an intersection operation on 2000 cases for each language setting, we get these consistently activated neurons across all the language settings: [6642, 15], [1648, 10], [4531, 8], [5077, 16] and [1392, 7], where the first coordinate represents the neuron index $\in [0,11007]$ and the second coordinate represents the layer $\in [0,31]$. We hope these data can accelerate the understanding and exploration of LLM-based MT.

\section{Scaling Experiments on LLaMA2-13B}\label{scaling_exps}
The scaling experiments on LLaMA2-13B are shown in Table \ref{tab:scaling_main_text}. The \textit{MT-I neurons} is the intersection of the top 100 MT neurons of all language settings (similar refinement like \textit{RPN-I}). We further include COMET22KIWI~\cite{rei-etal-2022-cometkiwi}, which is a reference-free evaluation method for a comprehensive evaluation of the translation quality. We can observe similar results for improving the language mismatch error compared with table \ref{tab:overall_main_text} on LLaMA2-7B. This means our detected Machine Translation heads exist and work in LLMs, and both heads and vectors matter for MT (the \textbf{MTV-I-D} achieves the best performance on improving the language mismatch error). Besides, our distributed MT head intervention method constantly improves the language mismatch issue in all language settings, which shows our finding is general enough for the MT task. Additionally, we also find that the KN method is not stable for improving the repetition error. We think there are two possible reasons. The first reason is the repetition error is much more complicated based on previous findings~ \cite{repe1}. The second reason is the KN theory is over-simplified to solve it~ \cite{oversimple-kn}.

\section{Case-Editing on Repetition Error}
We do the case-editing experiment on repetition cases only. Table~\ref{tab:case_editing} shows the results. Even though we hope to see some general KN set, KN effectively changes a token rather than a region of errors. In most cases, changing a repeated token can prevent the repetition error, which indicates that the repetition behavior has a connection with some token patterns. We leave this part with current research on repetition~ \cite{degene-report}as our future works.

\begin{table*}[h]
\vspace{-0.1in}
\centering
\resizebox{\textwidth}{!}{
\begin{tabular}{lcccccccccccccccc}
\toprule
\textbf{\textit{}}&  \multicolumn{4}{c}{\textbf{de$\rightarrow$en}} & \multicolumn{4}{c}{\textbf{en$\rightarrow$de}} & \multicolumn{4}{c}{\textbf{zh$\rightarrow$en}} & \multicolumn{4}{c}{\textbf{en$\rightarrow$zh}}  \\
\midrule
\midrule
\textbf{\textit{Zero-Shot}} & \textbf{L}($\downarrow$)  & \textbf{B}($\uparrow$)  &\textbf{C-1}($\uparrow$)  &\textbf{C-2}($\uparrow$)  &\textbf{L}($\downarrow$)  &\textbf{B}($\uparrow$)  &\textbf{C-1}($\uparrow$)  &\textbf{C-2}($\uparrow$)  &\textbf{L}($\downarrow$)  & \textbf{B}($\uparrow$)  & \textbf{C-1}($\uparrow$)  & \textbf{C-2}($\uparrow$) & \textbf{L}($\downarrow$)  & \textbf{B}($\uparrow$) & \textbf{C-1}($\uparrow$)  & \textbf{C-2}($\uparrow$)\\

\midrule                     
llama2-13b &$0.0073$ &$38.9229$ &$0.7949$ &$0.7796$ &$0.1472$ &$23.5531$ &$0.7116$ &$0.6911$ &$0.1514$ &$22.3365$ &$0.7244$ &$0.6882$ &$0.0304$ &$19.5333$ &$0.7401$ &$0.7578$ \\
+\textit{MTV} &$0.0\%$ &$-1.01\%$ &$0.1\%$ &$-0.13\%$ &$7.34\%$ &$-0.02\%$ &$-0.8\%$ &$-0.97\%$ &$-8.92\%$ &$-12.86\%$ &$-1.49\%$ &$-1.95\%$ &$16.45\%$ &$0.1\%$ &$-0.05\%$ &$-0.01\%$ \\
+\textit{MTV-I} &$0.0\%$ &$-3.96\%$ &$-0.03\%$ &$-0.53\%$ &$-78.06\%$ &$7.31\%$ &$2.26\%$ &$4.24\%$ &$-51.25\%$ &$-17.4\%$ &$-0.25\%$ &$0.93\%$ &$-35.2\%$ &$-5.29\%$ &$-0.16\%$ &$-0.28\%$ \\
+\textit{MTV-I-D} &\underline{$-50.68\%$} &$-1.4\%$ &$0.06\%$ &$-0.15\%$ &\underline{$-81.73\%$} &$12.75\%$ &$5.27\%$ &$7.18\%$ &\underline{$-59.84\%$} &$1.48\%$ &$2.07\%$ &$3.4\%$ &\underline{$-46.71\%$} &$-9.0\%$ &$0.14\%$ &$0.42\%$ \\
+\textit{MT neurons} &$0.0\%$ &$-0.46\%$ &$-0.14\%$ &$0.06\%$ &$-1.22\%$ &$1.96\%$ &$-0.42\%$ &$-0.39\%$ &$1.92\%$ &$-0.98\%$ &$-0.36\%$ &$-0.35\%$ &$-5.26\%$ &$0.88\%$ &$0.18\%$ &$0.25\%$ \\
+\textit{MT-I neurons} &$0.0\%$ &$-0.01\%$ &$-0.04\%$ &$-0.06\%$ &$-2.45\%$ &$1.51\%$ &$0.04\%$ &$-0.07\%$ &$1.25\%$ &$4.95\%$ &$0.14\%$ &$-0.1\%$ &$-8.55\%$ &$2.41\%$ &$-0.03\%$ &$0.03\%$ \\

\midrule
\textbf{\textit{One-Shot}} & \textbf{R}($\downarrow$)  & \textbf{B}($\uparrow$)  &\textbf{C-1}($\uparrow$)  &\textbf{C-2}($\uparrow$)  &\textbf{R}($\downarrow$)  &\textbf{B}($\uparrow$)  &\textbf{C-1}($\uparrow$)  &\textbf{C-2}($\uparrow$)  &\textbf{R}($\downarrow$)  & \textbf{B}($\uparrow$)  & \textbf{C-1}($\uparrow$)  & \textbf{C-2}($\uparrow$) & \textbf{R}($\downarrow$)  & \textbf{B}($\uparrow$) & \textbf{C-1}($\uparrow$)  & \textbf{C-2}($\uparrow$)\\

\midrule
llama2-13b &$0.0007$ &$39.9956$ &$0.801$ &$0.7849$ &$0.0087$ &$30.0092$ &$0.761$ &$0.7591$ &$0.0128$ &$30.7585$ &$0.7638$ &$0.74$ &$0.0009$ &$21.5548$ &$0.7513$ &$0.7709$ \\
+\textit{MT neurons} &\underline{$-100.0\%$} &$0.13\%$ &$-0.01\%$ &$0.06\%$ &\underline{$-17.24\%$} &$-0.16\%$ &$-0.24\%$ &$-0.05\%$ &$-12.5\%$ &$0.18\%$ &$-0.04\%$ &$-0.11\%$ &$0.0\%$ &$1.03\%$ &$0.05\%$ &$0.14\%$ \\
+\textit{MT-I neurons} &$-42.86\%$ &$-0.01\%$ &$0.02\%$ &$-0.04\%$ &$-3.45\%$ &$0.26\%$ &$0.08\%$ &$-0.09\%$ &\underline{$-35.16\%$} &$3.23\%$ &$0.16\%$ &$0.08\%$ &$55.56\%$ &$0.13\%$ &$-0.07\%$ &$-0.1\%$ \\
+\textit{RPN} &$-42.86\%$ &$-0.23\%$ &$-0.04\%$ &$-0.06\%$ &\underline{$-17.24\%$} &$0.32\%$ &$-0.01\%$ &$0.01\%$ &$-3.91\%$ &$-1.06\%$ &$-0.3\%$ &$-0.34\%$ &\underline{$-11.11\%$} &$0.41\%$ &$-0.03\%$ &$-0.03\%$ \\
+\textit{RPN-I} &\underline{$-100.0\%$} &$0.25\%$ &$0.02\%$ &$0.04\%$ &$-13.79\%$ &$-0.19\%$ &$0.05\%$ &$-0.03\%$ &$0.78\%$ &$-1.7\%$ &$-0.35\%$ &$-0.43\%$ &$55.56\%$ &$-0.63\%$ &$-0.27\%$ &$-0.26\%$ \\

\bottomrule
\end{tabular}
}
\vspace{-0.1in}
\caption{
Scaling experiments on LLaMA2-13B on $\mathcal{D}_{test}$ under all language settings. For evaluation metrics: \textbf{L}: Language mismatch ratio; \textbf{R}: Repetition ratio; \textbf{B}: BLEU; \textbf{C-1}: COMET22DA, \textbf{C-2}: COMET22KIWI, where \textbf{B}, \textbf{C-1} and \textbf{C-2} evaluate the general translation quality. Other notations and abbreviations follow Table~\ref{tab:direct_adaptation_main_text} for detailed methods, where \textbf{-I} means the intersection part based on all language settings. We \underline{underline} the best performance for improving the corresponding errors.
}
\vspace{-0.1in}
\label{tab:scaling_main_text}
\end{table*}

\begin{table}[!h]
\centering
\resizebox{0.48\textwidth}{!}{
\begin{tabular}{l|ll}
\hline
\textbf{Methods}     & \textbf{MTV-I-D} & \textbf{RPN-I}     \\ \hline
Per-token (\textit{original}) & 0.037s & 0.036s \\ \hline
Per-token (\textit{edited})   & 0.042s & 0.044s \\ \hline
Per-case (\textit{original})  & 1.62s  & 1.33s  \\ \hline
Per-case (\textit{edited})    & 2.22s  & 1.70s  \\ \hline
\end{tabular}}
\caption{The computation cost of our proposed methods compared with no-editing on 1000 cases in the zh$\rightarrow$en setting. The \textit{original} means using the LLaMA2-7B model only without any modifications. The \textit{edited} means using our proposed MTV-I-D or RPN-I editing. Other notations and abbreviations are following Table~\ref{tab:direct_adaptation_main_text}}
\label{computation_cost}
\end{table}
\section{Other Discussions}
\label{other_discussion}
\begin{table}[!ht]
\centering
\resizebox{0.48\textwidth}{!}{
\begin{tabular}{l|ccccc}
\hline
Template   & MMLU   & TruthfulQA & MMLU-Pro & CMMLU  & CommonQA \\ \hline
LLaMA2-7B   & 0.4593 & 0.3897     & 0.1860   & 0.3267 & 0.5659        \\ \hline
+\textit{MTV-I-D}& \textbf{0.4699} & 0.3884     & \textbf{0.1975}   & 0.3253 & \textbf{0.5684}        \\ \hline
+\textit{RPN-I} & 0.4613 & \textbf{0.3898}     & 0.1862   & 0.3254 & 0.5659        \\ \hline
\end{tabular}}
\caption{Evaluate our proposed methods on general LLM benchmarks. The results are averaged accuracy for corresponding benchmarks.}
\label{llm_benchmark_test}
\end{table}

\paragraph{The effect on inherent abilities of LLMs} One consideration when applying model editing methods to LLMs is their effect on other LLM abilities \cite{rome, emergence}. Even though we focus on machine translation and extracting corresponding vital components, these components may also be responsible for other potential abilities of LLMs like ICL. To further explore the effect of our proposed methods on other abilities of LLMs, we evaluate the models patched (i.e., edited) by our proposed methods on five popular and representative LLM benchmarks of general abilities. The evluation Benchmarks include MMLU~ \cite{benchmarkmmlu} (testing the general multitask abilities of LLM), TruthfulQA~ \cite{benchmarktruthfulqa} (testing the truthfulness of LLM, where we use the difficult multi-true task named mc2), MMLU-Pro~ \cite{benchmarkmmlupro} (a more discriminative benchmark compared to MMLU), CMMLU~ \cite{benchmarkcmmlu} (testing the effect of methods on Non-English languages, i.e., Chinese) and CommonsenseQA~ \cite{benchmarkcommonsense} (testing the commonsense knowledge of LLM).
We take the averaged accuracy measure, following the original papers. Higher values indicate better performance. Table~\ref{llm_benchmark_test} shows nearly no sharp performance drop on these evaluation benchmarks. This suggests that our editing methods do not affect other abilities of LLMs after patching MT-related components. One surprising phenomenon is that sometimes our methods even help improve the performance, as evidenced by the bold values, which could be explained by the enhanced language understanding abilities after editing (For implementation details, please see Appendix~\ref{app_implement_details}).

\paragraph{Additional computation cost.} One advantage of our proposed methods compared to traditional fine-tuning or LoRA is they incur minimal and acceptable computational costs. We do a concrete analysis for the time complexity:
For conventional Transformers, the computational complexity for a complete forward pass is $\mathcal{O}((N^2d+Nd^2)\cdot L)$, where $N,d,L$ refer to the input length, hidden dimension and the number of transformer layers. \textbf{For the Function Vectors (FV) based editing methods}, we finally select 12 attention heads to manipulate, resulting in an additional computation overhead of $\mathcal{O}(12\cdot \frac{d}{H})=\mathcal{O}(\frac{d}{H})$ (head addition takes the $\mathcal{O}(\frac{d}{H})$ complexity, where $H$ is the total number of heads of each layer). In practice, we directly manipulate the head output (after multiplication with the attention output matrix $W_O\in \mathbb{R}^{\frac{d}{H}\times d}$), making the complexity $\mathcal{O}(\frac{d^2}{H})$. 
\textbf{For the Knowledge Neurons (KN) based editing methods}, we select 31 neurons to manipulate, resulting in an additional computational overhead of $\mathcal{O}(31\cdot d')=\mathcal{O}(d')$, where $d'$ is the up-projection hidden dimension in the middle the MLP module. For LLaMA2, $d'=\frac{8}{3}d$. In practice, we manipulate the MLP output (after multiplication with the down-projection matrix $W_{down}\in \mathbb{R}^{d'\times d}$), making the complexity $\mathcal{O}(d'\cdot d)=\mathcal{O}(\frac{8d^2}{3})$. An optimised implementation can reduce the complexity of the FV editing and the KN editing to $\mathcal{O}(\frac{d}{H})$ and $\mathcal{O}(\frac{8d}{3})$ by only manipulating a single row of the multiplied matrices.

Apart from the theoretical analysis, we run an empirical experiment in the zh$\rightarrow$en setting, averaging over 1,000 cases. We calculate the statistical results per token and per case. Table~\ref{computation_cost} shows that the additional computational overhead brought by our editing methods is marginal.

\end{document}